\documentclass[11pt, prepint]{article}

\usepackage[preprint]{acl}

\usepackage{times}
\usepackage{latexsym}

\usepackage[T1]{fontenc}

\usepackage[utf8]{inputenc}

\usepackage{microtype}

\usepackage{inconsolata}

\usepackage{multirow}
\usepackage{graphicx}
\usepackage{amsmath} 
\usepackage{booktabs}
\usepackage{makecell}
\usepackage{xcolor}
\usepackage{algorithm}
\usepackage{algpseudocode}
\usepackage{tcolorbox}
\tcbuselibrary{breakable, skins}
\definecolor{darkblue}{RGB}{15, 15, 165}
\hypersetup{
    colorlinks=true,    
    citecolor=darkblue, 
    linkcolor=darkblue,
}
\title{WaveFilter: Enhancing the Long-Context Capability of Diffusion LLMs via Wavelet-Guided KV Cache Filtering}

\author{
 \textbf{Jinnan Yang\textsuperscript{1,4}},
 \textbf{Yan Wang\textsuperscript{2}},
 \textbf{Zhen Bi\textsuperscript{3}},
 \textbf{Kehao Wu\textsuperscript{1}},
\\
 \textbf{Xiaojie Li\textsuperscript{1}},
 \textbf{Jungang Lou\textsuperscript{3}},
 \textbf{Zechao Li\textsuperscript{1\text{\dag}}},
 \textbf{Jing Liu\textsuperscript{4\text{\dag}}}
\\
\textsuperscript{1}Nanjing University of Science and Technology,
\\
\textsuperscript{2}Alibaba Group,
\textsuperscript{3}Huzhou Normal University,
\\
\textsuperscript{4}Institute of Automation, Chinese Academy of Sciences
\\
\text{\textsuperscript{\dag}Corresponding authors}
}

\begin{document}
\maketitle
\begin{abstract}
Diffusion Large Language Models (DLMs) have demonstrated significant advantages across various tasks. However, constrained by their multi-step iterative inference mechanism, their computational overhead and inference latency in long-context tasks have become core bottlenecks restricting their large-scale deployment. When processing long sequences, existing Key-Value (KV) caching mechanisms often face a dilemma where generation quality degrades drastically, where the core challenge lies in precisely and efficiently filtering critical tokens within ultra-long contexts. Inspired by the human reading process, we propose \textbf{WaveFilter}, a universal and training-free caching framework. This framework innovatively introduces the wavelet transform for decomposition of long sequences to achieve precise identification of key tokens, based on which a sparse KV Cache is constructed to compute the final contextual representation. Experimental results demonstrate that WaveFilter, as a plug-and-play generic framework, significantly enhances the performance of existing mainstream KV Cache methods in complex long-context tasks.
\end{abstract}

\section{Introduction}
Owing to their non-autoregressive nature and bidirectional contextual modeling capability, diffusion large language models (DLMs) \cite{Nie2025} have demonstrated unique advantages in tasks such as text-to-image generation, dialogue systems, and code generation \cite{Sahoo2024,Gupta2024,Gong2025}. However, constrained by their multi-step iterative inference mechanism, DLMs incur substantially higher computational complexity and inference latency than autoregressive models \cite{Li2022,Li2025}. This heavy computational burden has become a core bottleneck limiting their large-scale deployment. To alleviate the overhead caused by repeated computation, researchers have introduced Key-Value (KV) caching mechanisms into DLMs \cite{Ma2025}. By caching the Key and Value vectors from previous steps, this method enables subsequent generation to directly reuse prior computation results. This effectively avoids redundant computation over already generated context, thereby reducing inference latency and improving generation efficiency.

\begin{figure}[t]
\centering
\includegraphics[width=0.5\textwidth]{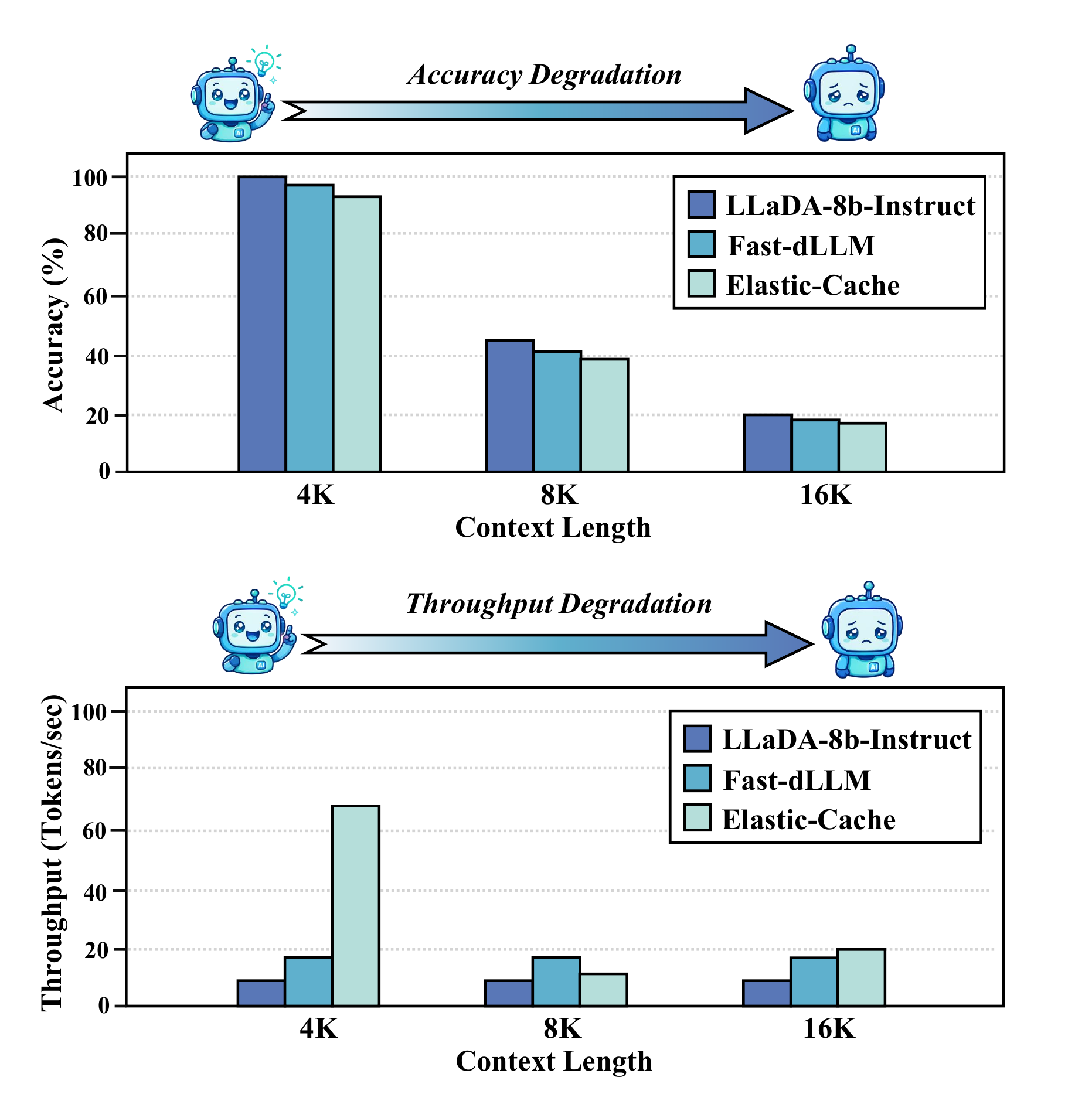} 
\caption{Performance comparison at different context lengths on the \textbf{niah\_single\_1} subset of \textbf{Ruler}. (a) illustrates the \textbf{accuracy (\%)} of the \textbf{LLaDA-8b-Instruct} and its variants with various KV Cache methods; (b) displays the corresponding \textbf{throughput (Tokens/sec)}.}
\label{fig:1}
\end{figure}

However, directly extending KV caching mechanisms to DLMs for handling complex long-context tasks still faces significant challenges. On the one hand, as shown in Figure~\ref{fig:1}a, the performance of LLaDA-8B-Instruct sharply declines as input length increases, indicating that the model struggles to maintain generation robustness in long-context tasks \cite{Liu2026}. On the other hand, existing KV caching mechanisms have been insufficiently studied in the context of DLMs applied to complex long-context tasks. As illustrated in Figure~\ref{fig:1}a and Figure~\ref{fig:1}b, although Fast-dLLM \cite{wu2025} and Elastic-Cache \cite{Tri2025} can provide certain acceleration benefits for short-text tasks, their throughput rapidly deteriorates as context length grows, often accompanied by further reductions in accuracy. \textbf{The core challenge lies in the extreme difficulty of precisely identifying and filtering tokens that make critical contributions to the denoising process within ultra-long context sequences.} Therefore, developing a universal, plug-and-play enhancement framework to empower existing caching methods, such as Fast-dLLM and Elastic-Cache, to extend efficiently and robustly to long-context tasks remains a imperative issue.

To accurately screen for crucial tokens in long-context tasks, we draw inspiration from the human cognitive habit of "skimming before scanning." Humans typically skim the entire text first to rapidly construct a macro-level contextual semantic structure, and subsequently perform localized scanning targeted at specific questions to locate and extract key information, ultimately achieving efficient and precise question answering. Inspired by this cognitive process, we propose \textbf{WaveFilter}, a universal and training-free framework. The core of this framework lies in the introduction of the \textbf{Discrete Wavelet Transform (DWT) for KV cache compression}. By reducing cache length and filtering out high-frequency noise while fully preserving time-domain information, WaveFilter successfully facilitates the rapid construction of macro contextual semantic structures. Following this, a multi-scale recursive filtering mechanism is employed to simulate localized scanning, precisely pinning down the tokens most relevant to the question to achieve highly accurate question answering.

Specifically, at the initial time step, the DWT is first utilized to extract the semantic features of the cache, and the attention mechanism is employed to identify the initial critical tokens targeted by the query vector. Based on this, multi-scale recursive filtering is performed on the initial important tokens to determine the final critical tokens. These final tokens are directly utilized to dynamically construct a sparse KV Cache, which subsequently participates in attention computation with the current query vector to precisely forge the final context representation. In summary, the primary contributions of this paper are as follows:

\begin{itemize}
    \item This paper proposes WaveFilter, a universal and training-free KV caching framework for long-context tasks. By mimicking human cognitive reading habits, this framework seamlessly empowers existing KV Cache methods, effectively resolving their performance degradation in long-context tasks.
    \item WaveFilter innovatively introduces the wavelet transform to achieve multi-scale token filtering: by compressing the cache drastically while preserving critical information, it precisely identifies relevant regions at a negligible cost, thereby effectively resolving the challenge of identifying pivotal tokens within massive caches.
    \item Experimental results demonstrate that WaveFilter, as a universal plug-and-play framework, can be seamlessly integrated into various existing KV caching strategies: while maintaining competitive generation speed, it boosts model performance in complex long-context tasks, consistently outperforming standalone KV caching methods.
\end{itemize}

\section{Preliminary}
\subsection{Key-Value Cache in Masked Diffusion Models}
\begin{figure*}
    \centering
    \includegraphics[width=\textwidth]{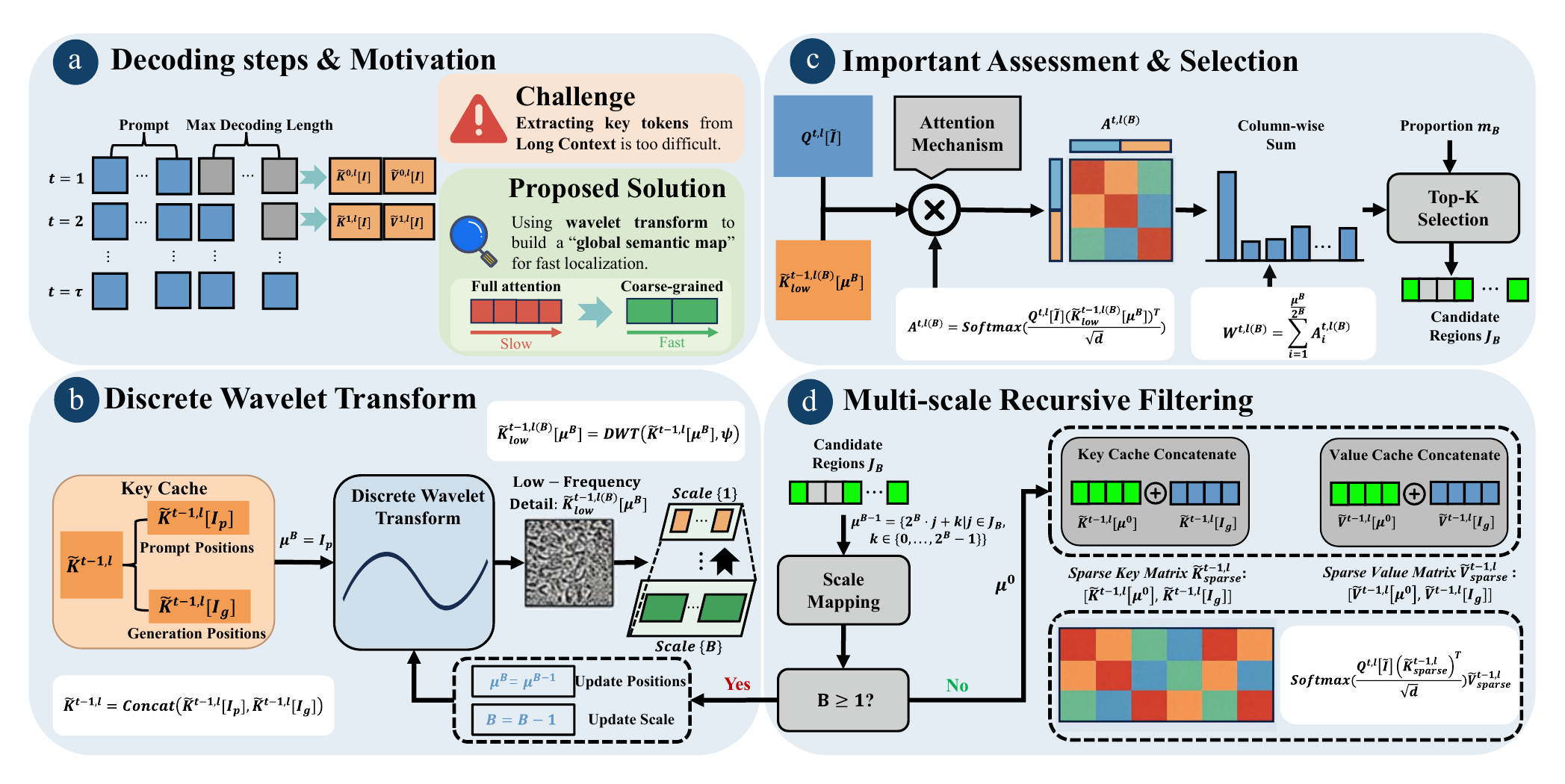} 
    \caption{Schematic pipeline of WaveFilter. Consists of four parts: (a) \textbf{Decoding steps \& Motivation}: It illustrates the decoding mechanism of discrete diffusion models and introduces a coarse-to-fine retrieval strategy to address the challenge of key token extraction in long-context tasks. (b) \textbf{Discrete Wavelet Transform}: Decomposition is performed on the cache keys via DWT to extract low-frequency components. (c) \textbf{Importance Assessment \& Selection}: Computes the correlation between the query vector and the low-frequency components of the cached keys, and utilizes Top-K selection to identify key candidate regions. (d) \textbf{Multi-scale Recursive Filtering}: Recursively refines candidate regions across different scales, ultimately selecting the most informative sparse KV matrix.}
    \label{fig:2}
\end{figure*}

Masked Diffusion Models (MDMs) replace traditional continuous noise addition with random masking and iterative blank-filling, enabling the parallel generation of discrete data \cite{Dickstein2015,Austin2021,Campbell2022,Sahoo2024_1}. To optimize generation efficiency during the reverse process, the KV Cache mechanism is integrated into its Transformer backbone. As illustrated in Figure~\ref{fig:2}a, at the initial timestep $t$ (where $t=1$), the model performs full computation over all positions $I=\{1,2,\dots,N\}$. At the $l$-th layer, the current hidden state $h^{1,l}$ is projected into query vectors
$Q^{1,l}[I]$, key vectors $K^{1,l}[I]$, and value vectors $V^{1,l}[I]$ via learnable projection matrices $W_Q^{1,l}$, $W_K^{1,l}$, and $W_V^{1,l}$. The attention output and the corresponding initialization of the KV Cache for this layer are formulated as:

\begin{equation}
\small
A^{1,l}[I] = \text{Softmax}\left(\frac{Q^{1,l}[I](K^{1,l}[I])^T}{\sqrt{d}}\right)V^{1,l}[I].
\label{eq:1}
\end{equation}

Subsequently, the KV pairs computed at the initial step are saved to the cache, with the initialization formally defined as:

\begin{equation}
\small
\left\{
\begin{aligned}
    \widetilde{K}^{1,l}[I] &= K^{1,l}[I] \\
    \widetilde{V}^{1,l}[I] &= V^{1,l}[I]
\end{aligned}
\right..
\label{eq:2}
\end{equation}

In subsequent timesteps $t > 1$, the model performs inference only for the set of generation positions $\widetilde{I}$. By reusing the cached keys $\widetilde{K}$ and values $\widetilde{V}$ stored from previous timesteps, the attention computation is simplified to:

\begin{equation}
\small
A^{t,l}[\widetilde{I}] = \text{Softmax}\left(\frac{Q^{t,l}[\widetilde{I}](\widetilde{K}^{t-1,l}[I])^T}{\sqrt{d}}\right)\widetilde{V}^{t-1,l}[I].
\label{eq:3}
\end{equation}

Subsequently, the cache is dynamically updated using the KV pairs computed at the current step:
\begin{equation}
\small
\left \{
\begin{aligned}
\widetilde{K}^{t,l}[\widetilde{I}] = K^{t,l}[\widetilde{I}] \\ \widetilde{V}^{t,l}[\widetilde{I}] = V^{t,l}[\widetilde{I}])
\end{aligned}
\right..
\label{eq:4}
\end{equation}

Inference based on the KV cache significantly reduces computational complexity and inference latency during the reverse process. Through this mechanism, the model substantially enhances the efficiency of discrete sequence generation while maintaining the global context modeling capabilities of the Transformer.

\subsection{Discrete Wavelet Transform}
The Discrete Wavelet Transform (DWT) is a time-frequency analysis method used for signal decomposition \cite{Yao2022,Kiruluta2025}. DWT decomposes a signal $x[n]$ into low-frequency approximation components and high-frequency detail components through a pair of complementary filter banks. A single-level decomposition process can be formulated as:
\begin{equation}
\small
\left \{
\begin{aligned}
A_1[n] = \sum_{k} x[k] \cdot g[2n-k]  \\ 
D_1[n] = \sum_{k} x[k] \cdot h[2n-k]
\end{aligned}
\right..
\label{eq:5}
\end{equation}

where $g[n]$ and $h[n]$ denote the low-pass and high-pass filter coefficients, respectively, and the subscript $2n$ represents the downsampling operation. The core advantage of DWT lies in its recursive nature: following the first level of decomposition, the approximation coefficients $A_1$ can serve as the input for the subsequent level of the filter bank. This iterative process constructs a multi-level pyramid structure. After $L$ levels of decomposition, the original signal is represented by the set of components $\{A_L, D_L, D_{L-1}, \dots, D_1\}$. In the context of long-sequence modeling, $A_L$ captures the global semantic information of the signal, while the various detail components $D_j$ preserve local fluctuations across different resolutions.

\section{Methodology}
\label{sec:Method}
This section provides a detailed exposition of the core mechanisms of WaveFilter framework. To address the challenge of identifying and filtering critical tokens in long-context tasks, this paper proposes a KV Cache framework based on the DWT. As shown in Figure~\ref{fig:2}, the overall algorithmic pipeline consists of two distinct stages (refer to Appendix~\ref{appendix:Algorithm} for the detailed algorithm). The first stage, \textbf{coarse-grained global perception (Section~\ref{sec:Coarse-grained})} employs the DWT to construct low-frequency components, enabling the rapid localization of semantic regions that contain critical information within a compressed space. The second stage, \textbf{fine-grained local localization (Section~\ref{sec:Fine-grained})}, which achieves precise extraction of important tokens through multi-scale recursive filtering, based on which a sparse KV Cache is constructed. This coarse-to-fine strategy not only effectively maintains speed competitiveness but also significantly boosts the overall performance of the model.

\subsection{Coarse-grained Global Perception: Constructing Global Semantic Outlines}
\label{sec:Coarse-grained}
For long-sequence processing, the standard attention mechanism faces severe computational challenges when extracting key tokens highly relevant to the query from extensive contexts, primarily due to its $O(N^2)$ time complexity. To alleviate this bottleneck, this section proposes a novel method for constructing a "Global Semantic Map" based on the DWT, aiming to rapidly and accurately localize potential regions harboring critical tokens by leveraging semantic distributions.

Let $Q^{t,l}[\widetilde{I}] \in {R}^{\widetilde{I} \times d}$ denote the query vector at layer $l$ and time step $t \; (t>1)$. The cached key vectors comprise prompt positions $I_p$ and generation positions $I_g$. Specifically, ${\widetilde{K}}^{t-1,l}[I_p] \in {R}^{I_p \times d}$ and ${\widetilde{K}}^{t-1,l}[I_g] \in {R}^{I_g \times d}$ represent the cached key vectors at layer $l$ and time step $t-1$ for the prompt and generation positions, respectively. To capture sequential features across different receptive fields, as illustrated in Figure~\ref{fig:2}b, we first extract the low-frequency components of ${\widetilde{K}}^{t-1,l}[I_p] \in {R}^{I_p \times d}$ via the DWT with a wavelet basis $\psi$:

\begin{equation}
\small
\widetilde{K}_{low}^{t-1,l(B)}[I_p] = DWT(\widetilde{K}^{t-1,l}[I_p], \psi),
\label{eq:6}
\end{equation}

where $\widetilde{K}_{low}^{t-1,l(B)}[I_p] \in {R}^{\frac{I_p}{2^B} \times d}$ denotes the low-frequency component at scale $B$. By filtering out high-frequency noise, these components effectively capture the semantic outlines of the prompt long sequences at a coarse-grained level. As shown in Figure~\ref{fig:2}c, since the sequence dimension is significantly compressed at scale $B$, we can compute the perception weight matrix of $Q^{t,l}[\widetilde{I}]$ relative to $\widetilde{K}_{low}^{t-1,l(B)}[I_p]$ via the attention mechanism at a negligible computational cost:

\begin{equation}
\small
A^{t,l(B)} = Softmax\left(\frac{Q^{t,l}[\widetilde{I}] (\widetilde{K}_{low}^{t-1,l(B)}[I_p])^T}{\sqrt{d}}\right),
\label{eq:7}
\end{equation}

where $A^{t,l(B)} \in {R}^{\widetilde{I} \times \frac{I_p}{2^B}}$ characterizes the correlations between tokens within the compressed space. To identify the candidate regions in $\widetilde{K}_{low}^{t-1,l(B)}[I_p]$ with the highest semantic relevance to $Q^{t,l}[\widetilde{I}]$, we perform a column-wise summation of $A^{t,l(B)}$ to obtain the importance evaluation vector:

\begin{equation}
\small
W^{t,l(B)} = \sum_{i=1}^{\frac{I_p}{2^B}} A_i^{t,l(B)},
\label{eq:8}
\end{equation}

$W^{t,l(B)}$ precisely characterizes the contribution of different tokens in $\widetilde{K}_{low}^{t-1,l(B)}[I_p]$ to the current query from a macroscopic perspective. Finally, based on the importance scores, we select a proportion $m_B$ of the most significant regions to determine the candidate region set $J_B$:
\begin{equation}
\small
J_B = \underset{j \in \{1, \dots, \frac{I_p}{2^B}\}}{Top-K} (W_j^{t, l(B)}, m_B).
\label{eq:9}
\end{equation}

Through the aforementioned process, we successfully implement a coarse-grained semantic macro-screening stage for long-context sequences. By leveraging the compression characteristics of the wavelet transform to filter out redundant information without pursuing localized granular precision, this stage rapidly outlines the critical semantic intervals of $Q^{t,l}[\widetilde{I}]$ over $\widetilde{K}_{low}^{t-1,l(B)}[I_p]$. This drastically shrinks the subsequent search space, thereby guiding the fine-grained screening in Section~\ref{sec:Fine-grained} with an exceptionally low computational overhead.

\subsection{Fine-grained Local Localization: Multi-scale Recursive Filtering}
\label{sec:Fine-grained}
Although the initial candidate set $J_B$ effectively narrows the search space, these regions inevitably contain tokens irrelevant to the current query. Relying solely on these coarse regions for KV Cache updates and reuse decisions would compromise generation accuracy. To address this, we construct a multi-scale recursive filtering method, as illustrated in Figure~\ref{fig:2}d. First, the frequency-domain regions $J_B$ at scale $B$ are mapped back to the index space of the prompt sequence:

\begin{equation}
\small
\mu^{B-1} = \{ 2^B \cdot j + k \mid j \in J_B, k \in \{0, \dots, 2^B-1\} \},
\label{eq:10}
\end{equation}

Where $\mu^{B-1}$ denotes the set of indices covered by the candidate regions in the prompt sequence at scale $B$. Subsequently, the corresponding subsequence tokens are extracted from ${\widetilde{K}}^{t-1,l}[I_p]$ via a tensor indexing operation to construct a sparse key matrix, which is then subjected to a wavelet transform to obtain the higher-resolution low-frequency approximation component (at scale $B-1$):

\begin{equation}
\small
\widetilde{K}_{low}^{t-1,l(B-1)}[\mu^{B-1}] = DWT(\widetilde{K}^{t-1,l}[\mu^{B-1}],\psi).
\label{eq:11}
\end{equation}

Compared to $\widetilde{K}_{low}^{t-1,l(B)}[I_p]$, the component $\widetilde{K}_{low}^{t-1,l(B-1)}[\mu^{B-1}]$ offers higher resolution and finer local representations, enabling more granular refinement of the candidate region localization. By repeating the procedure described in Figure~\ref{fig:2}c, we derive a more precise candidate set $J_{B-1}$.

By iteratively executing the aforementioned evaluation and selection process, important regions are recursively refined layer by layer from scale $B$ to scale 1. This hierarchical filtering mechanism enables cross-scale alignment from macroscopic semantic perception to precise token localization, thereby obtaining the most informative token index $\mu^0$ from ${\widetilde{K}}^{t-1,l}[I_p]$. Finally, the corresponding sparse key matrix $\widetilde{K}^{t-1,l}[\mu^0]$ and value matrix $\widetilde{V}^{t-1,l}[\mu^0]$ are extracted via a tensor index selection operation. They are then concatenated with $\widetilde{K}^{t-1,l}[I_g]$ and $\widetilde{V}^{t-1,l}[I_g]$ of the current time step along the sequence dimension, respectively, to construct the final sparse key matrix ${\widetilde{K}}_{sparse}^{t-1,l}$ and value matrix ${\widetilde{V}}_{sparse}^{t-1,l}$. These matrices are then utilized alongside the query vector of the current step to compute the final contextual representation.

\section{Experiments}
\subsection{Experiments Setup}

\paragraph{Implementation Details.} All experiments are conducted on a \textbf{single NVIDIA A800 80GB GPU}. We evaluate WaveFilter using \textbf{LLaDA-8b-Instruct} \cite{Nie2025} and \textbf{Dream-v0-Base-7B} \cite{Ye2025} across a diverse set of benchmarks, including \textbf{Longbench} \cite{Bai2024} and \textbf{Ruler} \cite{Hsieh2024}. Detailed hyperparameter configurations are provided in Appendix~\ref{appendix:Experiment Setup}. To ensure a fair comparison, we re-run LLaDA-8b-Instruct \cite{Nie2025}, Dream-v0-Base-7B \cite{Ye2025}, Fast-dLLM \cite{wu2025}, and Elastic-Cache \cite{Tri2025} under identical hardware and software environments. \textbf{Evaluation Framework and Metrics}. We utilize the \textbf{lm-eval-harness} \cite{Leo2024} framework for our evaluation. Following the protocol established by Fast-dLLM and Elastic-Cache, \textbf{throughput} is measured as the average tokens per second (Tokens/sec) calculated until the generation process terminates.
\paragraph{Confidence-Aware Decoding.}  We implement the confidence-aware decoding strategy from FastdLLM \cite{wu2025}. Unlike the fixed-step unmasking mechanism in baseline Diffusion LLMs, this strategy performs dynamic filtering by introducing a threshold $\epsilon$: tokens are selected only when their confidence scores exceed this limit. This mechanism allows the model to adaptively adjust its decoding scope per iteration based on prediction quality, thereby significantly enhancing inference efficiency. Consequently, our experiments focus on evaluating the additional acceleration gains specifically provided by KV caching under this unified decoding framework.

\subsection{Performance and Efficiency Evaluation}
To evaluate the effectiveness and efficiency of the proposed WaveFilter framework, we conduct benchmark testing on the LongBench and Ruler datasets. We integrate WaveFilter into several competitive baselines and perform a detailed experimental comparison. Table~\ref{tab:1} and Table~\ref{tab:2} present the accuracy, throughput, and relative speedup ratios of each baseline. Additionally, Appendix~\ref{appendix:Performance} details the total generated tokens and the total runtime for each baseline across all datasets.

\begin{table*}[t]
\caption{Comprehensive benchmark results of \textbf{LLaDA-8B-Instruct} and \textbf{Dream-v0-Base-7B} on \textbf{Longbench}. Each cell displays the accuracy (top row), the decoding throughput in Tokens/sec, and the speedup ratio relative to the LLaDA and Dream baselines (bottom row, \textcolor{blue}{blue: throughput}/\textcolor{orange}{orange: speedup}). The symbol "-" denotes \textbf{out of memory} errors.}
\label{tab:1}
\centering
\footnotesize 
\setlength{\tabcolsep}{3pt} 
\renewcommand{\arraystretch}{1.3} 
\resizebox{\textwidth}{!}{
\begin{tabular}{l cc ccc cc cc c}
\toprule
 & \multicolumn{2}{c}{\textbf{Single-Doc QA}} & \multicolumn{3}{c}{\textbf{Multi-Doc QA}} & \multicolumn{2}{c}{\textbf{Few-shot Learning}} & \multicolumn{2}{c}{\textbf{Synthetic}} & \multicolumn{1}{c}{\textbf{Code}} \\
\cmidrule(lr){2-3} \cmidrule(lr){4-6} \cmidrule(lr){7-8} \cmidrule(lr){9-10} \cmidrule(lr){11-11} 
& Qsp & MulF & HQA & 2WQA & MSQ & TREC & TrQA & PsgC & PsgR & LCC \\
\midrule
\textit{LLaDA-8B-Instruct} 
& \makecell{24.64 \\ \scriptsize \textcolor{blue}{2.56}(\textcolor{orange}{$1.00\times$})} 
& \makecell{24.52 \\ \scriptsize \textcolor{blue}{2.03}(\textcolor{orange}{$1.00\times$})} 
& \makecell{3.36 \\ \scriptsize \textcolor{blue}{1.18}(\textcolor{orange}{$1.00\times$})} 
& \makecell{3.95 \\ \scriptsize \textcolor{blue}{2.00}(\textcolor{orange}{$1.00\times$})} 
& \makecell{1.08 \\ \scriptsize \textcolor{blue}{0.94}(\textcolor{orange}{$1.00\times$})}
& \makecell{33.17 \\ \scriptsize \textcolor{blue}{0.23}(\textcolor{orange}{$1.00\times$})}
& \makecell{67.33 \\ \scriptsize \textcolor{blue}{0.16}(\textcolor{orange}{$1.00\times$})}
& \makecell{1.10 \\ \scriptsize \textcolor{blue}{1.84}(\textcolor{orange}{$1.00\times$})} 
& \makecell{15.89 \\ \scriptsize \textcolor{blue}{2.47}(\textcolor{orange}{$1.00\times$})} 
& \makecell{61.66 \\ \scriptsize \textcolor{blue}{6.75}(\textcolor{orange}{$1.00\times$})}
\\

+ Fast-dLLM 
& \makecell{22.03 \\ \scriptsize \textcolor{blue}{15.01}(\textcolor{orange}{$5.86\times$})} 
& \makecell{24.25 \\ \scriptsize \textcolor{blue}{11.59}(\textcolor{orange}{$5.71\times$})} 
& \makecell{2.64 \\ \scriptsize \textcolor{blue}{8.35}(\textcolor{orange}{$7.08\times$})} 
& \makecell{3.69 \\ \scriptsize \textcolor{blue}{12.87}(\textcolor{orange}{$6.44\times$})} 
& \makecell{0.92 \\ \scriptsize \textcolor{blue}{7.33}(\textcolor{orange}{$7.80\times$})}
& \makecell{33.03 \\ \scriptsize \textcolor{blue}{0.89}(\textcolor{orange}{$3.87\times$})}
& \makecell{67.13 \\ \scriptsize \textcolor{blue}{1.31}(\textcolor{orange}{$8.19\times$})} 
& \makecell{1.03 \\ \scriptsize \textcolor{blue}{8.30}(\textcolor{orange}{$4.51\times$})} 
& \makecell{20.04 \\ \scriptsize \textcolor{blue}{9.27}(\textcolor{orange}{$3.75\times$})} 
& \makecell{60.05 \\ \scriptsize \textcolor{blue}{27.27}(\textcolor{orange}{$4.04\times$})} 
\\

+ Fast-dLLM \& WaveFilter
& \makecell{\textbf{28.73} \\ \scriptsize \colorbox{gray!20}{\textcolor{blue}{6.80}(\textcolor{orange}{$2.66\times$})}} 
& \makecell{\textbf{26.37} \\ \scriptsize \colorbox{gray!20}{\textcolor{blue}{7.42}(\textcolor{orange}{$3.66\times$})}} 
& \makecell{\textbf{2.65} \\ \scriptsize \colorbox{gray!20}{\textcolor{blue}{5.71}(\textcolor{orange}{$4.84\times$})}} 
& \makecell{\textbf{4.95} \\ \scriptsize \colorbox{gray!20}{\textcolor{blue}{8.66}(\textcolor{orange}{$4.33\times$})}} 
& \makecell{\textbf{1.02} \\ \scriptsize \colorbox{gray!20}{\textcolor{blue}{5.37}(\textcolor{orange}{$5.71\times$})}}
& \makecell{\textbf{31.10} \\ \scriptsize \colorbox{gray!20}{\textcolor{blue}{0.52}(\textcolor{orange}{$2.26\times$})}}
& \makecell{\textbf{64.99} \\ \scriptsize \colorbox{gray!20}{\textcolor{blue}{0.86}(\textcolor{orange}{$5.38\times$})}}
& \makecell{\textbf{1.90} \\ \scriptsize \colorbox{gray!20}{\textcolor{blue}{5.27}(\textcolor{orange}{$2.86\times$})}}
& \makecell{\textbf{22.32} \\ \scriptsize \colorbox{gray!20}{\textcolor{blue}{6.13}(\textcolor{orange}{$2.48\times$})}}
& \makecell{\textbf{60.08} \\ \scriptsize \colorbox{gray!20}{\textcolor{blue}{19.44}(\textcolor{orange}{$2.88\times$})}}
\\

+ Elastic-Cache 
& \makecell{- \\ -} 
& \makecell{21.34 \\ \scriptsize \textcolor{blue}{10.42}(\textcolor{orange}{$5.13\times$})} 
& \makecell{- \\ -} 
& \makecell{- \\ -} 
& \makecell{- \\ -}
& \makecell{32.78 \\ \scriptsize \textcolor{blue}{0.58}(\textcolor{orange}{$2.52\times$})}
& \makecell{- \\ -}
& \makecell{- \\ -} 
& \makecell{21.56 \\ \scriptsize \textcolor{blue}{5.32}(\textcolor{orange}{$2.15\times$})}  
& \makecell{- \\ -}

\\
+  Elastic-Cache \& WaveFilter
& \makecell{- \\ -}
& \makecell{\textbf{24.38} \\ \scriptsize \colorbox{gray!20}{\textcolor{blue}{10.88}(\textcolor{orange}{$5.36\times$})}}
& \makecell{- \\ -}
& \makecell{- \\ -}
& \makecell{- \\ -}
& \makecell{\textbf{32.86} \\ \scriptsize \colorbox{gray!20}{\textcolor{blue}{0.73}(\textcolor{orange}{$3.17\times$})}}
& \makecell{- \\ -}
& \makecell{- \\ -}
& \makecell{\textbf{21.94} \\ \scriptsize \colorbox{gray!20}{\textcolor{blue}{5.11}(\textcolor{orange}{$2.07\times$})}}
& \makecell{- \\ -}

\\
\midrule
\textit{Dream-v0-Base-7B} 
& \makecell{24.68 \\ \scriptsize \textcolor{blue}{3.47}(\textcolor{orange}{$1.00\times$})} 
& \makecell{37.43 \\ \scriptsize \textcolor{blue}{2.21}(\textcolor{orange}{$1.00\times$})} 
& \makecell{12.41 \\ \scriptsize \textcolor{blue}{1.10}(\textcolor{orange}{$1.00\times$})} 
& \makecell{5.66 \\ \scriptsize \textcolor{blue}{2.29}(\textcolor{orange}{$1.00\times$})} 
& \makecell{5.46 \\ \scriptsize \textcolor{blue}{0.86}(\textcolor{orange}{$1.00\times$})} 
& \makecell{70.00 \\ \scriptsize \textcolor{blue}{3.23}(\textcolor{orange}{$1.00\times$})}
& \makecell{87.90 \\ \scriptsize \textcolor{blue}{1.37}(\textcolor{orange}{$1.00\times$})} 
& \makecell{0.80 \\ \scriptsize \textcolor{blue}{2.20}(\textcolor{orange}{$1.00\times$})} 
& \makecell{14.43 \\ \scriptsize \textcolor{blue}{2.87}(\textcolor{orange}{$1.00\times$})} 
& \makecell{19.30 \\ \scriptsize \textcolor{blue}{9.41}(\textcolor{orange}{$1.00\times$})}
\\

+ Fast-dLLM 
& \makecell{24.58 \\ \scriptsize \textcolor{blue}{28.05}(\textcolor{orange}{$8.08\times$})} 
& \makecell{36.60 \\ \scriptsize \textcolor{blue}{21.40}(\textcolor{orange}{$9.68\times$})} 
& \makecell{11.85 \\ \scriptsize \textcolor{blue}{12.83}(\textcolor{orange}{$11.66\times$})} 
& \makecell{4.68 \\ \scriptsize \textcolor{blue}{21.97}(\textcolor{orange}{$9.59\times$})} 
& \makecell{3.76 \\ \scriptsize \textcolor{blue}{10.69}(\textcolor{orange}{$12.43\times$})}
& \makecell{70.00 \\ \scriptsize \textcolor{blue}{25.16}(\textcolor{orange}{$12.43\times$})}
& \makecell{89.29 \\ \scriptsize \textcolor{blue}{13.71}(\textcolor{orange}{$10.01\times$})} 
& \makecell{1.05 \\ \scriptsize \textcolor{blue}{14.08}(\textcolor{orange}{$6.40\times$})} 
& \makecell{16.69 \\ \scriptsize \textcolor{blue}{16.71}(\textcolor{orange}{$5.82\times$})} 
& \makecell{15.01 \\ \scriptsize \textcolor{blue}{45.17}(\textcolor{orange}{$4.80\times$})}
\\

+ Fast-dLLM \& WaveFilter
& \makecell{\textbf{27.95} \\ \scriptsize \colorbox{gray!20}{\textcolor{blue}{16.66}(\textcolor{orange}{$4.80\times$})}} 
& \makecell{\textbf{37.93} \\ \scriptsize \colorbox{gray!20}{\textcolor{blue}{11.38}(\textcolor{orange}{$5.15\times$})}} 
& \makecell{\textbf{18.13} \\ \scriptsize \colorbox{gray!20}{\textcolor{blue}{7.55}(\textcolor{orange}{$6.86\times$})}}
& \makecell{\textbf{5.83} \\ \scriptsize \colorbox{gray!20}{\textcolor{blue}{13.34}(\textcolor{orange}{$5.83\times$})}}
& \makecell{\textbf{10.29} \\ \scriptsize \colorbox{gray!20}{\textcolor{blue}{6.06}(\textcolor{orange}{$7.05\times$})}} 
& \makecell{\textbf{69.50} \\ \scriptsize \colorbox{gray!20}{\textcolor{blue}{12.78}(\textcolor{orange}{$3.96\times$})}} 
& \makecell{\textbf{89.16} \\ \scriptsize \colorbox{gray!20}{\textcolor{blue}{7.42}(\textcolor{orange}{$5.42\times$})}}
& \makecell{\textbf{1.32} \\ \scriptsize \colorbox{gray!20}{\textcolor{blue}{8.32}(\textcolor{orange}{$3.78\times$})}}
& \makecell{\textbf{20.06} \\ \scriptsize \colorbox{gray!20}{\textcolor{blue}{10.08}(\textcolor{orange}{$3.51\times$})}}
& \makecell{\textbf{13.81} \\ \scriptsize \colorbox{gray!20}{\textcolor{blue}{28.67}(\textcolor{orange}{$3.05\times$})}}
\\

+ Elastic-Cache 
& \makecell{- \\ -} 
& \makecell{32.85 \\ \scriptsize \textcolor{blue}{31.34}(\textcolor{orange}{$14.18\times$})}
& \makecell{- \\ -} 
& \makecell{4.46 \\ \scriptsize \textcolor{blue}{13.96}(\textcolor{orange}{$6.10\times$})} 
& \makecell{- \\ -} 
& \makecell{35.54 \\ \scriptsize \textcolor{blue}{10.53}(\textcolor{orange}{$6.10\times$})} 
& \makecell{- \\ -}
& \makecell{- \\ -}
& \makecell{7.14 \\ \scriptsize \textcolor{blue}{13.73}(\textcolor{orange}{$4.78\times$})} 
& \makecell{- \\ -} 
\\

+ Elastic-Cache \& WaveFilter
& \makecell{- \\ -} 
& \makecell{\textbf{39.64} \\ \scriptsize \colorbox{gray!20}{\textcolor{blue}{37.77}(\textcolor{orange}{$17.09\times$})}} 
& \makecell{- \\ -}
& \makecell{\textbf{5.89} \\ \scriptsize \colorbox{gray!20}{\textcolor{blue}{12.31}(\textcolor{orange}{$5.38\times$})}}
& \makecell{- \\ -} 
& \makecell{\textbf{33.49} \\ \scriptsize \colorbox{gray!20}{\textcolor{blue}{11.02}(\textcolor{orange}{$3.41\times$})}} 
& \makecell{- \\ -}
& \makecell{- \\ -}
& \makecell{\textbf{16.77} \\ \scriptsize \colorbox{gray!20}{\textcolor{blue}{12.98}(\textcolor{orange}{$4.52\times$})}}
& \makecell{- \\ -}
\\
\bottomrule
\end{tabular}
}
\end{table*}

\paragraph{LongBench Results.} Table~\ref{tab:1} summarizes the experimental results across various long-context tasks, including single-document QA, multi-document QA, few-shot learning, synthetic tasks, and code generation. The results indicate that introducing the WaveFilter framework as a plug-in component into existing cache management methods significantly boosts model accuracy on long-context tasks at the cost of a marginal efficiency drop. For instance, in single-document QA, multi-document QA, synthetic tasks, and code generation, \textbf{Fast-dLLM \& WaveFilter} consistently improves the accuracy of LLaDA-8B-Instruct compared to the vanilla Fast-dLLM. Meanwhile, \textbf{Elastic-Cache \& WaveFilter} not only enhances the accuracy of LLaDA-8B-Instruct over Elastic-Cache but also improves throughput and effectively reduces the total execution time on the MulF dataset. Furthermore, when extended to the Dream-v0-Base-7B, WaveFilter similarly yields significant improvements in experimental accuracy.

\paragraph{Ruler Results.} To further evaluate WaveFilter framework under varying context lengths, we report the performance on the Ruler benchmark at 4K and 8K context windows in Table~\ref{tab:2}. The empirical findings reveal that as the context length increases, standard cache management methods like Fast-dLLM and Elastic-Cache suffer from severe performance degradation compared to the dense LLaDA-8B-Instruct model. However, integrating the WaveFilter framework successfully mitigates this accuracy drop for both Fast-dLLM and Elastic-Cache. For example, on the vt dataset, the dense LLaDA-8B-Instruct model achieves an accuracy of 40.6\%. When utilizing Fast-dLLM and Elastic-Cache, the accuracy plunges to 19.08\% and 25.72\%, respectively. In contrast, the incorporation of the WaveFilter framework recovers the accuracy to 21.56\% and 26.04\%, thereby effectively alleviating the performance deterioration.

In summary, the consistent results across both LongBench and Ruler benchmarks demonstrate the exceptional effectiveness and generalization capability of the WaveFilter framework. As a plug-and-play, universal component, WaveFilter not only significantly alleviates the accuracy attenuation inherent in KV Cache methods during long-context tasks but also maintains or even enhances throughput on specific benchmarks. These findings validate the rationale of preserving critical contextual information through multi-scale filtering mechanisms.

\begin{table*}
\caption{Comprehensive benchmark results of \textbf{LLaDA-8B-Instruct} on \textbf{Ruler}. Each cell displays the accuracy (top row), the decoding throughput in Tokens/sec, and the speedup ratio relative to the LLaDA baselines (bottom row, \textcolor{blue}{blue: throughput}/\textcolor{orange}{orange: speedup}).}
\label{tab:2}
\centering
\footnotesize
\resizebox{\textwidth}{!}{
\begin{tabular}{l | cccccccccc | c}
\toprule
 & s1 & s2 & m1 & m2 & mv & mq & vt & cwe & qa1 & qa2 & Context Length \\
\midrule
\textit{LLaDA-8B-Instruct} 
& \makecell{100.00 \\ \scriptsize \textcolor{blue}{3.80}(\textcolor{orange}{$1.00\times$})} 
& \makecell{100.00 \\ \scriptsize \textcolor{blue}{8.12}(\textcolor{orange}{$1.00\times$})} 
& \makecell{100.00 \\ \scriptsize \textcolor{blue}{7.33}(\textcolor{orange}{$1.00\times$})}
& \makecell{92.40 \\ \scriptsize \textcolor{blue}{4.99}(\textcolor{orange}{$1.00\times$})}
& \makecell{100.00 \\ \scriptsize \textcolor{blue}{4.61}(\textcolor{orange}{$1.00\times$})}
& \makecell{99.80 \\ \scriptsize \textcolor{blue}{6.35}(\textcolor{orange}{$1.00\times$})}
& \makecell{96.92 \\ \scriptsize \textcolor{blue}{6.43}(\textcolor{orange}{$1.00\times$})}
& \makecell{30.56 \\ \scriptsize \textcolor{blue}{6.35}(\textcolor{orange}{$1.00\times$})}
& \makecell{79.18 \\ \scriptsize \textcolor{blue}{4.25}(\textcolor{orange}{$1.00\times$})}
& \makecell{78.80 \\ \scriptsize \textcolor{blue}{5.25}(\textcolor{orange}{$1.00\times$})}
& 4K \\
+ Fast-dLLM 
& \makecell{99.60 \\ \scriptsize \textcolor{blue}{22.46}(\textcolor{orange}{$5.91\times$})} 
& \makecell{100.00 \\ \scriptsize \textcolor{blue}{21.10}(\textcolor{orange}{$2.60\times$})} 
& \makecell{100.00 \\ \scriptsize \textcolor{blue}{21.40}(\textcolor{orange}{$2.92\times$})} 
& \makecell{88.00 \\ \scriptsize \textcolor{blue}{22.95}(\textcolor{orange}{$4.60\times$})}
& \makecell{97.55 \\ \scriptsize \textcolor{blue}{29.60}(\textcolor{orange}{$6.42\times$})}
& \makecell{99.55 \\ \scriptsize \textcolor{blue}{31.43}(\textcolor{orange}{$4.95\times$})} 
& \makecell{93.08 \\ \scriptsize \textcolor{blue}{28.43}(\textcolor{orange}{$4.42\times$})} 
& \makecell{1.58 \\ \scriptsize \textcolor{blue}{30.42}(\textcolor{orange}{$4.79\times$})}
& \makecell{77.32 \\ \scriptsize \textcolor{blue}{12.08}(\textcolor{orange}{$2.84\times$})}
& \makecell{77.40 \\ \scriptsize \textcolor{blue}{17.39}(\textcolor{orange}{$3.31\times$})}
& 4K
\\
+ Fast-dLLM  \& WaveFilter
& \makecell{\textbf{99.80} \\ \scriptsize \colorbox{gray!20}{\textcolor{blue}{5.95}(\textcolor{orange}{$1.57\times$})}}
& \makecell{\textbf{100.00} \\ \scriptsize \colorbox{gray!20}{\textcolor{blue}{6.15}(\textcolor{orange}{$0.76\times$})}}
& \makecell{\textbf{100.00} \\ \scriptsize \colorbox{gray!20}{\textcolor{blue}{5.81}(\textcolor{orange}{$0.79\times$})}}
& \makecell{\textbf{85.80} \\ \scriptsize \colorbox{gray!20}{\textcolor{blue}{13.91}(\textcolor{orange}{$2.79\times$})}}
& \makecell{\textbf{98.20} \\ \scriptsize \colorbox{gray!20}{\textcolor{blue}{18.62}(\textcolor{orange}{$4.04\times$})}}
& \makecell{\textbf{98.60} \\ \scriptsize \colorbox{gray!20}{\textcolor{blue}{17.65}(\textcolor{orange}{$2.78\times$})}}
& \makecell{\textbf{94.00} \\ \scriptsize \colorbox{gray!20}{\textcolor{blue}{21.48}(\textcolor{orange}{$3.34\times$})}}
& \makecell{\textbf{1.64} \\ \scriptsize \colorbox{gray!20}{\textcolor{blue}{16.16}(\textcolor{orange}{$2.54\times$})}}
& \makecell{\textbf{77.35} \\ \scriptsize \colorbox{gray!20}{\textcolor{blue}{6.58}(\textcolor{orange}{$1.55\times$})}}
& \makecell{\textbf{76.80} \\ \scriptsize \colorbox{gray!20}{\textcolor{blue}{11.92}(\textcolor{orange}{$2.27\times$})}}
& 4K
\\
+ Elastic-Cache 
& \makecell{99.20 \\ \scriptsize \textcolor{blue}{65.28}(\textcolor{orange}{$17.18\times$})} 
& \makecell{100.00 \\ \scriptsize \textcolor{blue}{21.43}(\textcolor{orange}{$2.64\times$})}
& \makecell{100.00 \\ \scriptsize \textcolor{blue}{29.74}(\textcolor{orange}{$4.06\times$})}
& \makecell{90.20 \\ \scriptsize \textcolor{blue}{17.39}(\textcolor{orange}{$3.48\times$})}
& \makecell{96.85 \\ \scriptsize \textcolor{blue}{18.75}(\textcolor{orange}{$4.07\times$})}
& \makecell{99.75 \\ \scriptsize \textcolor{blue}{24.72}(\textcolor{orange}{$3.89\times$})}
& \makecell{96.48 \\ \scriptsize \textcolor{blue}{22.21}(\textcolor{orange}{$3.45\times$})}
& \makecell{16.50 \\ \scriptsize \textcolor{blue}{9.03}(\textcolor{orange}{$1.42\times$})}
& \makecell{80.37 \\ \scriptsize \textcolor{blue}{92.51}(\textcolor{orange}{$21.77\times$})}
& \makecell{77.20 \\ \scriptsize \textcolor{blue}{45.77}(\textcolor{orange}{$8.72\times$})}
& 4K
\\
+ Elastic-Cache \& WaveFilter
& \makecell{\textbf{99.60} \\ \scriptsize \colorbox{gray!20}{\textcolor{blue}{54.41}(\textcolor{orange}{$14.32\times$})}}
& \makecell{\textbf{100.00} \\ \scriptsize \colorbox{gray!20}{\textcolor{blue}{17.86}(\textcolor{orange}{$2.20\times$})}}
& \makecell{\textbf{100.00} \\ \scriptsize \colorbox{gray!20}{\textcolor{blue}{28.50}(\textcolor{orange}{$3.89\times$})}}
& \makecell{\textbf{87.40} \\ \scriptsize \colorbox{gray!20}{\textcolor{blue}{15.11}(\textcolor{orange}{$3.03\times$})}}
& \makecell{\textbf{97.15} \\ \scriptsize \colorbox{gray!20}{\textcolor{blue}{17.01}(\textcolor{orange}{$3.69\times$})}}
& \makecell{\textbf{99.78} \\ \scriptsize \colorbox{gray!20}{\textcolor{blue}{20.49}(\textcolor{orange}{$3.23\times$})}}
& \makecell{\textbf{96.78} \\ \scriptsize \colorbox{gray!20}{\textcolor{blue}{19.05}(\textcolor{orange}{$2.96\times$})}}
& \makecell{\textbf{16.95} \\ \scriptsize \colorbox{gray!20}{\textcolor{blue}{7.44}(\textcolor{orange}{$1.17\times$})}}
& \makecell{\textbf{80.75} \\ \scriptsize \colorbox{gray!20}{\textcolor{blue}{88.14}(\textcolor{orange}{$4.88\times$})}}
& \makecell{\textbf{77.80} \\ \scriptsize \colorbox{gray!20}{\textcolor{blue}{43.22}(\textcolor{orange}{$8.23\times$})}}

& 4K
\\
\midrule
\textit{LLaDA-8B-Instruct} 
& \makecell{57.00 \\ \scriptsize \textcolor{blue}{4.08}(\textcolor{orange}{$1.00\times$})} 
& \makecell{76.40 \\ \scriptsize \textcolor{blue}{1.73}(\textcolor{orange}{$1.00\times$})}
& \makecell{63.60 \\ \scriptsize \textcolor{blue}{2.08}(\textcolor{orange}{$1.00\times$})}
& \makecell{43.00 \\ \scriptsize \textcolor{blue}{1.91}(\textcolor{orange}{$1.00\times$})}
& \makecell{55.75 \\ \scriptsize \textcolor{blue}{3.08}(\textcolor{orange}{$1.00\times$})}
& \makecell{54.45 \\ \scriptsize \textcolor{blue}{2.71}(\textcolor{orange}{$1.00\times$})}
& \makecell{40.60 \\ \scriptsize \textcolor{blue}{3.14}(\textcolor{orange}{$1.00\times$})}
& \makecell{26.84 \\ \scriptsize \textcolor{blue}{2.81}(\textcolor{orange}{$1.00\times$})}
& \makecell{49.63 \\ \scriptsize \textcolor{blue}{2.65}(\textcolor{orange}{$1.00\times$})}
& \makecell{66.80 \\ \scriptsize \textcolor{blue}{2.39}(\textcolor{orange}{$1.00\times$})}
& 8K
\\
+ Fast-dLLM 
& \makecell{45.40 \\ \scriptsize \textcolor{blue}{17.06}(\textcolor{orange}{$4.18\times$})} 
& \makecell{55.20 \\ \scriptsize \textcolor{blue}{11.44}(\textcolor{orange}{$6.61\times$})} 
& \makecell{61.40 \\ \scriptsize \textcolor{blue}{12.90}(\textcolor{orange}{$6.20\times$})} 
& \makecell{40.60 \\ \scriptsize \textcolor{blue}{11.53}(\textcolor{orange}{$6.04\times$})}
& \makecell{43.80 \\ \scriptsize \textcolor{blue}{19.86}(\textcolor{orange}{$6.45\times$})}
& \makecell{52.10 \\ \scriptsize \textcolor{blue}{17.92}(\textcolor{orange}{$6.61\times$})}
& \makecell{19.08 \\ \scriptsize \textcolor{blue}{17.58}(\textcolor{orange}{$5.60\times$})}
& \makecell{4.54 \\ \scriptsize \textcolor{blue}{13.76}(\textcolor{orange}{$4.90\times$})}
& \makecell{47.87 \\ \scriptsize \textcolor{blue}{15.50}(\textcolor{orange}{$5.85\times$})}
& \makecell{63.00 \\ \scriptsize \textcolor{blue}{13.70}(\textcolor{orange}{$5.73\times$})}
& 8K
\\
+ Fast-dLLM \& WaveFilter
& \makecell{\textbf{47.80} \\ \scriptsize \colorbox{gray!20}{\textcolor{blue}{3.58}(\textcolor{orange}{$0.88\times$})}} 
& \makecell{\textbf{55.80} \\ \scriptsize \colorbox{gray!20}{\textcolor{blue}{7.68}(\textcolor{orange}{$4.44\times$})}}
& \makecell{\textbf{62.05} \\ \scriptsize \colorbox{gray!20}{\textcolor{blue}{8.65}(\textcolor{orange}{$4.16\times$})}}
& \makecell{\textbf{38.6} \\ \scriptsize \colorbox{gray!20}{\textcolor{blue}{7.05}(\textcolor{orange}{$3.69\times$})}}
& \makecell{\textbf{43.90} \\ \scriptsize \colorbox{gray!20}{\textcolor{blue}{13.16}(\textcolor{orange}{$4.27\times$})}}
& \makecell{\textbf{53.10} \\ \scriptsize \colorbox{gray!20}{\textcolor{blue}{12.16}(\textcolor{orange}{$4.49\times$})}}
& \makecell{\textbf{21.56} \\ \scriptsize \colorbox{gray!20}{\textcolor{blue}{13.11}(\textcolor{orange}{$4.18\times$})}}
& \makecell{\textbf{4.90} \\ \scriptsize \colorbox{gray!20}{\textcolor{blue}{9.70}(\textcolor{orange}{$3.45\times$})}}
& \makecell{\textbf{48.98} \\ \scriptsize \colorbox{gray!20}{\textcolor{blue}{10.70}(\textcolor{orange}{$4.04\times$})}}
& \makecell{\textbf{63.50} \\ \scriptsize \colorbox{gray!20}{\textcolor{blue}{9.21}(\textcolor{orange}{$3.85\times$})}}
& 8K
\\
+ Elastic-Cache 
& \makecell{44.80 \\ \scriptsize \textcolor{blue}{12.39}(\textcolor{orange}{$3.04\times$})} 
& \makecell{61.00 \\ \scriptsize \textcolor{blue}{7.01}(\textcolor{orange}{$4.05\times$})}
& \makecell{62.80 \\ \scriptsize \textcolor{blue}{7.91}(\textcolor{orange}{$3.80\times$})} 
& \makecell{42.00 \\ \scriptsize \textcolor{blue}{7.55}(\textcolor{orange}{$3.95\times$})}
& \makecell{50.60 \\ \scriptsize \textcolor{blue}{10.11}(\textcolor{orange}{$3.28\times$})}
& \makecell{56.00 \\ \scriptsize \textcolor{blue}{10.77}(\textcolor{orange}{$3.97\times$})}
& \makecell{25.72 \\ \scriptsize \textcolor{blue}{11.57}(\textcolor{orange}{$3.68\times$})}
& \makecell{7.48 \\ \scriptsize \textcolor{blue}{3.54}(\textcolor{orange}{$1.26\times$})}
& \makecell{47.87 \\ \scriptsize \textcolor{blue}{10.36}(\textcolor{orange}{$3.91\times$})}
& \makecell{63.40 \\ \scriptsize \textcolor{blue}{10.29}(\textcolor{orange}{$4.31\times$})}
& 8K
\\
+ Elastic-Cache \& WaveFilter
& \makecell{\textbf{45.20} \\ \scriptsize \colorbox{gray!20}{\textcolor{blue}{9.26}(\textcolor{orange}{$2.27\times$})}} 
& \makecell{\textbf{61.20} \\ \scriptsize \colorbox{gray!20}{\textcolor{blue}{7.43}(\textcolor{orange}{$4.29\times$})}} 
& \makecell{\textbf{63.20} \\ \scriptsize \colorbox{gray!20}{\textcolor{blue}{8.94}(\textcolor{orange}{$4.30\times$})}} 
& \makecell{\textbf{41.20} \\ \scriptsize \colorbox{gray!20}{\textcolor{blue}{5.62}(\textcolor{orange}{$2.79\times$})}} 
& \makecell{\textbf{51.12} \\ \scriptsize \colorbox{gray!20}{\textcolor{blue}{10.10}(\textcolor{orange}{$3.28\times$})}} 
& \makecell{\textbf{56.15} \\ \scriptsize \colorbox{gray!20}{\textcolor{blue}{10.08}(\textcolor{orange}{$3.72\times$})}} 
& \makecell{\textbf{26.04} \\ \scriptsize \colorbox{gray!20}{\textcolor{blue}{8.99}(\textcolor{orange}{$2.86\times$})}} 
& \makecell{\textbf{9.30} \\ \scriptsize \colorbox{gray!20}{\textcolor{blue}{2.38}(\textcolor{orange}{$0.85\times$})}} 
& \makecell{\textbf{48.85} \\ \scriptsize \colorbox{gray!20}{\textcolor{blue}{9.72}(\textcolor{orange}{$3.67\times$})}} 
& \makecell{\textbf{63.85} \\ \scriptsize \colorbox{gray!20}{\textcolor{blue}{8.40}(\textcolor{orange}{$3.51\times$})}} 
& 8K
\\
\midrule
\end{tabular}
}
\end{table*}

\subsection{Ablations and Analysis}
We conduct comprehensive ablation studies to evaluate the impact of two key hyperparameters: (1) \textbf{the recurrence scale $B$}, (2) \textbf{the candidate region proportion $m_B$}. This section aims to investigate how these hyperparameters influence both experimental accuracy and computational efficiency, thereby validating the optimality and rationale of our default configurations.
\subsubsection{Ablation and Analysis on Scale}
Figure~\ref{fig:3} illustrates the performance sensitivity of the WaveFilter framework with respect to $B$. The experimental results demonstrate that the choice of $B$ exerts a significant impact on both accuracy and computational efficiency.
\paragraph{Impact on Accuracy.} In terms of accuracy, a larger $B$ does not monotonically yield better performance. Experiments indicate that the model typically achieves peak accuracy at intermediate scales (e.g., $B=2$ or $B=3$). This suggests that an overly shallow scale ($B=1$) fails to effectively separate noise from key semantic features, whereas an excessively deep scale ($B=4$) may lead to the loss of critical semantic information.

\paragraph{Impact on Throughput and Total Runtime.} As $B$ gradually increases from 1 to 4, the throughput exhibits a clear downward trend, accompanied by an increase in total runtime. This is because higher values of $B$ entail more layers of recurrent filtering. Although a deeper hierarchy enables finer-grained feature alignment, each extra recurrence layer introduces extra computational overhead—specifically from wavelet transforms, perceptual weight matrices, and candidate region selection—which inevitably amplifies inference latency.

Based on the above analysis, we can conclude that the multi-scale recursive filtering described in Section~\ref{sec:Fine-grained} is essential. Furthermore, taking into account the trade-off between accuracy and computational efficiency, setting $B=2$ as the default configuration proves to be both optimal choice and highly reasonable in practice.

\subsubsection{Ablation and Analysis on Proportions}
\begin{figure*}[t]
    \centering
    \includegraphics[width=0.87\textwidth]{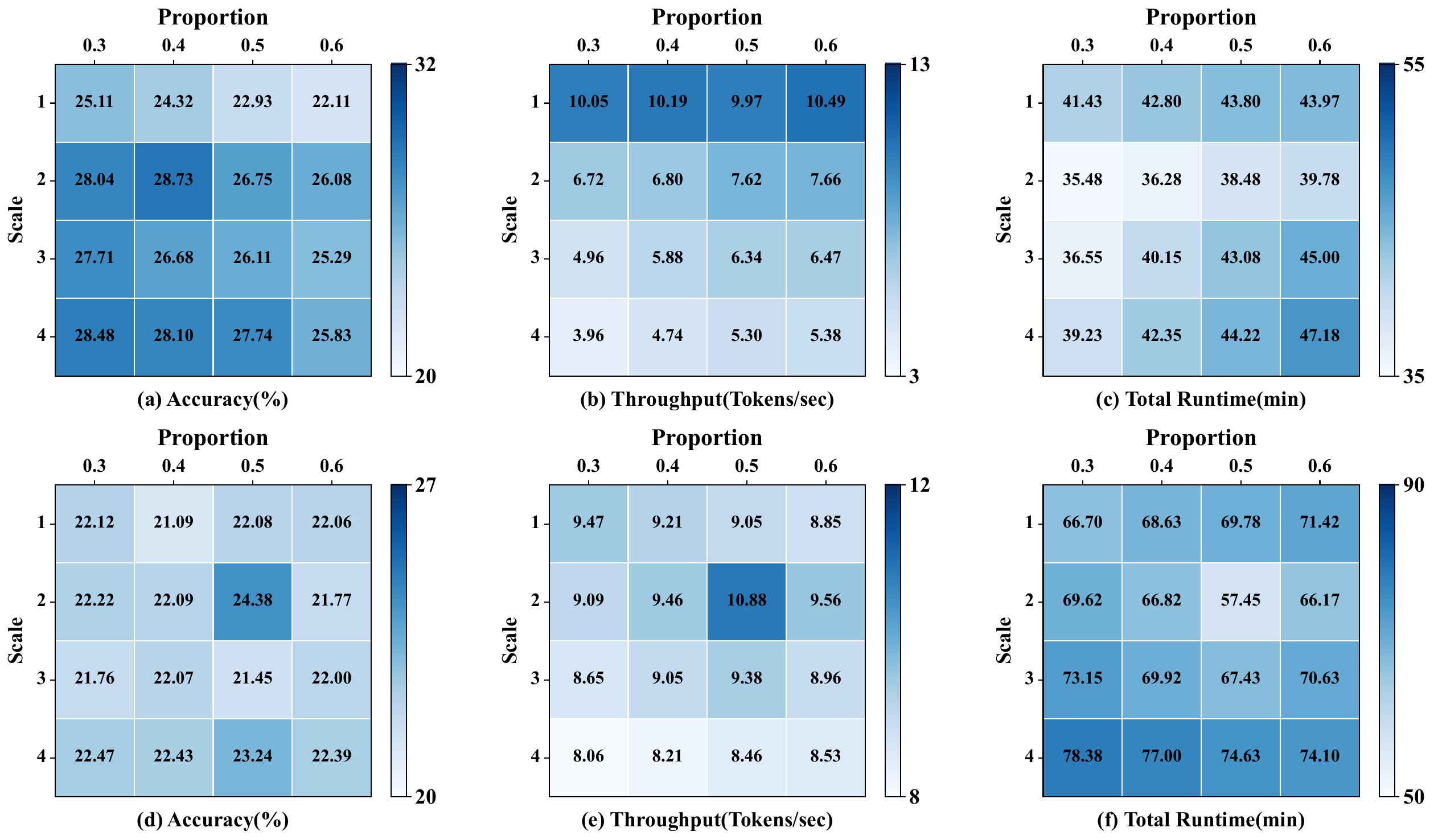} 
    \caption{Ablation study and performance analysis of \textbf{WaveFilter} combined with different caching methods on \textbf{LLaDA-8B-Instruct}. \textbf{(a)-(c)} evaluate \textbf{Fast-dLLM \& WaveFilter} on the \textbf{Longbench-Qsp} dataset, while \textbf{(d)-(f)} evaluate \textbf{Elastic-Cache \& WaveFilter} on the \textbf{Longbench-MulF} dataset. The heatmaps illustrate the joint effects of \textbf{Scale (y-axis)} and \textbf{Proportion (x-axis)} across three key metrics: \textbf{(a, d) Accuracy (\%)}, \textbf{(b, e) Throughput (Tokens/sec)}, and \textbf{(c, f) Total Runtime (min)}. Darker colors indicate higher values for each respective metric.}
    \label{fig:3}
\end{figure*}

Figure~\ref{fig:3} displays the performance sensitivity of WaveFilter framework relative to $m_B$. The parameter $m_B$ directly controls the retention proportion of tokens that proceed to subsequent screening after undergoing the wavelet transform. The empirical results reveal that the choice of $m_B$ significantly affects both accuracy and computational efficiency.

\paragraph{Impact on Accuracy.} As $m_B$ varies within a reasonable range of \textbf{[0.3, 0.6]}, the model suffers no performance degradation; instead, it exhibits a notable boost in accuracy. This clearly demonstrates that the WaveFilter framework can accurately identifies and preserves core semantic tokens that dictate generation quality, while efficiently discarding a vast amount of low-contribution, redundant tokens. Notably, moderate sparsification not only preserves foundational semantic integrity by keeping backbone tokens, but more importantly, it effectively mitigates noise interference within the attention mechanism by filtering out irrelevant tokens. This mechanism allows the model to focus more intensely on core contextual representations, thereby consistently enhancing experimental accuracy. This crucial finding fully substantiates the necessity and validity of executing deep sparse filtering on the KV Cache to balance accuracy and computational efficiency. 

\paragraph{Impact on Throughput and Total Runtime.} The impact of $m_B$ on computational efficiency exhibits a distinct non-linear characteristic. When $m_B$ is set to an overly small value, aggressive pruning expels an excessive number of tokens from the KV Cache. While this trims redundant information, it also discards part of the crucial semantics, which increases the model's inference time steps; consequently, generation throughput drops, and the total runtime increases. Conversely, when $m_B$ is too large, an influx of redundant tokens repopulates the KV Cache, and the heightened computational load similarly depresses throughput and prolongs runtime. The model maximizes generation throughput and substantially minimizes total runtime only when a reasonable selection proportion strikes the optimal balance between information integrity and sparsity—drastically reducing attention overhead without sacrificing core semantics.

Based on the above analysis, the selection proportion is paramount to the WaveFilter framework. Balancing both accuracy and computational efficiency, selecting an appropriate range of $m_B$ tailored to specific datasets represents the most effective and rational approach.

\section{Related Work}
\subsection{Diffusion Language Models}
Diffusion models initially achieved breakthroughs in continuous domain tasks, such as image and audio generation \cite{song2019,Ho2020,Dhariwal2021,Gupta2024}. Recently, to accommodate the discrete nature of text, researchers have introduced modeling approaches based on Markov, multinomial, and continuous-time frameworks, successfully extending this mechanism to the NLP domain \cite{Li2022,Gong2023,Lou2024}. Currently, the generation quality of masked diffusion models approaches that of autoregressive models \cite{Nie2025}, demonstrating competitive potential against models like LLaMA \cite{Grattafiori2024} and Qwen \cite{yang2024}. This not only provides an alternative to the autoregressive paradigm but has also expanded its impact into domains such as multimodality and code generation.

\subsection{LLM Acceleration Techniques}
Although LLM inference efficiency is constrained by quadratic computational overhead and memory bottlenecks in long sequences, it can be significantly accelerated via KV Cache by caching historical attention states \cite{zhang2023,li2024,Xiao2024,zhang2025,xiao2025}. In contrast, the caching mechanism in diffusion models is more complex; the presence of multiple denoising timesteps and distinct feature variations across steps severely diminishes caching effectiveness. Despite caching methods such as DeepCache \cite{Ma2024}, dLLM-Cache \cite{Liu2025}, dKV-Cache \cite{Ma2025}, Fast-dLLM \cite{wu2025}, Fast-dLLM v2 \cite{Wu2025_1}, Sparse-dLLM \cite{Song2026}, d\({}^{\mbox{2}}\)Cache \cite{Jiang2025}, and Elastic-Cache \cite{Tri2025}, they have yet to fully address the efficiency bottlenecks in long-context tasks.

\section{Conclusion}
To address the performance bottlenecks of existing diffusion model caching methods in long-context tasks, this paper proposes WaveFilter, a universal and training-free framework. This framework innovatively introduces the wavelet transform into the KV Cache, precisely locating highly correlated tokens within a reduced-dimensional space through multi-scale recursive filtering. Experimental results demonstrate that the WaveFilter framework effectively enhances the performance of existing caching methods in long-context tasks. Future work will further explore KV Cache optimization for DLMs in more complex tasks.

\newpage
\section*{Limitations}
While our proposed \textbf{WaveFilter framework} demonstrates significant efficacy, we acknowledge several limitations that warrant further investigation in future work.

First, relying solely on throughput to evaluate processing speed may lead to incomplete conclusions. As demonstrated by the empirical analyses in Tables~\ref{tab:1} and ~\ref{tab:2}, as well as Appendices~\ref{appendix:Performance} and ~\ref{append:example}, although the integration of WaveFilter results in a notable drop in throughput for the \textbf{s1} configuration within \textbf{Ruler}, it substantially shortens the total end-to-end execution time and delivers superior generation quality compared to the baseline. This phenomenon indicates that the conventional throughput metric (i.e., tokens per second) fails to fully capture the practical efficiency gains during inference. Therefore, a multi-dimensional evaluation that combines total runtime and output quality is essential for a fairer assessment of speed.

Second, despite the sequence compression, the perceptual weight computation still introduces non-negligible overhead. In Equation~\ref{eq:7}, while the wavelet transform effectively shortens the overall sequence length, computing the perceptual weight matrix via the attention mechanism inevitably increases computational complexity. This additional operational cost is the primary factor driving the decreased throughput and prolonged running times observed on certain specific datasets.

Lastly, a performance degradation occurs in text summarization tasks. As shown in the \textbf{few-shot learning} tasks in Table~\ref{tab:1}, the WaveFilter framework sparsifies the tokens corresponding to the prompt positions within the KV cache. Since complex summarization tasks heavily depend on both the global context and the specific instructions embedded within the prompt, this process of structural sparsification often inadvertently discards critical contextual prompts, thereby severely compromising the model's final generation performance on such highly specific tasks. 


\bibliography{references}

\newpage
\appendix
\section{Algorithm Procedure of WaveFilter}
\label{appendix:Algorithm}
To ensure the rigor and completeness of the proposed framework, the formal execution logic of the WaveFilter algorithm is detailed in Algorithm~\ref{alg:WaveFilter}. Following the theoretical foundations formulated in Section~\ref{sec:Method}, the execution procedure can be partitioned into three primary phases: \textbf{initialization, multi-resolution recursive filtering, and cache consolidation.} First, the algorithm initializes the complete sequence $x^0$ by combining the prompt sequence with positional placeholders for generation, and preserves the initial Key-Value (KV) cache across all Transformer layers (Lines 1–4). Second, it recursively processes the historical KV states via the Discrete Wavelet Transform (DWT), performing attention alignment and index screening across multiple resolutions to effectively extract the most informative sparse KV pairs (Lines 5–16). Finally, the isolated sparse cache is utilized to compute the final context representation for diffusion decoding (Lines 17–19).

\begin{algorithm*}[t]
\caption{The WaveFilter Algorithm}
\label{alg:WaveFilter}
\begin{algorithmic}[1]
\State \textbf{Input:} Prompt $x_{\text{prompt}}$, Generation Length $N$, Scale $B$, Wavelet $\psi$, Selection Proportions $\{m_b\}_{b=1}^B$, Threshold $\epsilon$.
\State \textbf{Initialize:} $x^0 \gets \{x_{\text{prompt}}; \text{[MASK]}, \dots, \text{[MASK]}\}$; $p \gets \text{length}(x_{\text{prompt}})$

\State $t \gets 1$; $I \gets \{1, \dots, p+N\}$; $\widetilde{I} \gets \{p+1, \dots, p+N\}$; $\widetilde{K}^{1,l}[I] \gets K^{1,l}[I]$; $\widetilde{V}^{1,l}[I] \gets V^{1,l}[I]$
\State $t \gets t+1$
\While{$\widetilde{I}^t \neq \emptyset$}
    \State $H^{t,1}[\widetilde{I}^t] \gets \text{Embedding}(x^t[\widetilde{I}^t])$
    \For{$l = 1, \dots, L$}
        \State $Q^{t,l}[\widetilde{I}^t], K^{t,l}[\widetilde{I}^t], V^{t,l}[\widetilde{I}^t] \gets \text{FFN}(H^{t,l}[\widetilde{I}^t])$
        \State $\mu^B \gets I_p$
        \While{$B \geq 1$}
            \State $\widetilde{K}_{low}^{t-1,l(B)}[\mu^B] \gets DWT(\widetilde{K}^{t-1,l}[\mu^B], \psi)$
            \State $A^{t,l(B)} \gets Softmax \left( (Q^{t,l}[\widetilde{I}^t] \cdot (\widetilde{K}_{low}^{t-1,l(B)}[\mu^B])^T) / \sqrt{d} \right)$
            \State $W^{t,l(B)} \gets \sum_{i=1}^{I_p/2^B} A_i^{t,l(B)}$
            \State $J_B \gets \mathop{Top-K}\limits_{j \in \{1, \dots, I_p/2^B\}} (W_j^{t,l(B)}, m_B)$
            \State $\mu^{B-1} \gets \{2^B \cdot j + k \mid j \in J_B, k \in \{0, \dots, 2^B-1\}\}$; $\mu^{B} \gets \mu^{B-1}$; $B \gets B-1$
        \EndWhile
        \State $\mu^0 \gets$ The most informative token indices from $\widetilde{K}^{t-1,l}[I_p]$
        \State ${\widetilde{K}}_{sparse}^{t-1,l} \gets [{\widetilde{K}}^{t-1,l}[\mu^0], {\widetilde{K}}^{t-1,l}[I_g]]$; ${\widetilde{K}}_{sparse}^{t-1,l} \gets [{\widetilde{V}}^{t-1,l}[\mu^0], {\widetilde{V}}^{t-1,l}[I_g]]$
        \State Compute the final context representation using $Q^{t,l}[\widetilde{I}^t]$, ${\widetilde{K}}_{sparse}^{t-1,l}$, and ${\widetilde{V}}_{sparse}^{t-1,l}$
    \EndFor
    \State Retain high-confidence tokens via threshold $\epsilon$, mask the rest; $t \gets t+1$
\EndWhile
\State \Return $x^{t-1}$
\end{algorithmic}
\end{algorithm*}

\section{Detailed Experiment Setup}
\label{appendix:Experiment Setup}
\subsection{Implementation Details}
We conduct all experiments on a \textbf{single NVIDIA A800 80GB GPU} to ensure a consistent hardware environment. Specifically, we implement our proposed \textbf{WaveFilter} framework on Fast-dLLM and Elastic-Cache, and evaluate its performance across two DLMs: \textbf{LLaDA-8b-Instruct} and \textbf{Dream-v0-Base-7B}. Our evaluation spans two long-context benchmarks: \textbf{Longbench} and \textbf{Ruler}. Within Longbench, we evaluate the following tasks: \textbf{single-document QA}, represented by Qasper \textbf{(Qsp)} and MultifieldQA\_en \textbf{(MulF)}; \textbf{multi-document QA}, including HotpotQA \textbf{(HQA)}, 2WikiMultihopQA \textbf{(2WQA)}, and Musique \textbf{(MSQ)}; the \textbf{few-shot learning} task \textbf{TREC} and \textbf{TriviaQA (TrQA)}; \textbf{synthetic tasks} PassageCount \textbf{(PsgC)} and PassageRetrieval\_en \textbf{(PsgR)}; and the \textbf{code completion} task \textbf{LCC}. For the Ruler benchmark, we further evaluate 10 core subtasks across four dimensions: (1) foundational and multi-target retrieval capabilities, covering single- to triple-needle retrieval (\textbf{niah\_single\_1 (s1), niah\_single\_2 (s2)}), multi-value retrieval (\textbf{niah\_multivalue (mv)}), and multi-query retrieval (\textbf{niah\_multiquery (mq)}); (2) 
multi-step logical reasoning and state-tracking abilities, evaluated through multi-key retrieval (\textbf{niah\_multikey\_1 (m1), niah\_multikey\_2 (m2)}) and variable tracking (\textbf{variable\_tracking (vt)}); (3) global information aggregation, measured via common word extraction (\textbf{cwe}); and (4) deep comprehension and question-answering performance on long texts, validated through single-hop QA (\textbf{qa\_squad (qa1)}) and multi-hop QA (\textbf{qa\_hotpot (qa2)}). To establish a rigorous and fair comparison, we re-evaluate all baseline methods, including the \textbf{confidence-based decoding} diffusion models \textbf{LLaDA-8B-Instruct} and \textbf{Dream-v0-Base-7B}, as well as the caching methods \textbf{Fast-dLLM} and \textbf{Elastic-Cache}. This procedure eliminates confounding variables arising from hardware or software discrepancies, ensuring that all observed performance variations are solely attributable to the methods themselves.

\subsection{Evaluation Framework and Metrics}
To ensure the standardization and reproducibility of our experiments, we utilize the \textbf{lm-eval-harness framework} to conduct all task-specific evaluations. We measure inference speed by throughput in tokens per second (\textbf{Tokens/sec}), which is calculated as the average number of tokens generated by the model over the entire sequence until an end-of-sequence (EOS) token is produced. Furthermore, our calculation methodology strictly aligns with those of \textbf{Fast-dLLM} and \textbf{Elastic-Cache} to ensure a rigorous and fair comparison of inference speed benchmarks across different methods.

\subsection{Hyperparameter Settings}
Table \ref{tab:hyperparameter} presents the hyperparameters used for \textbf{WaveFilter}. Specifically, for both the \textbf{LongBench} and \textbf{Ruler} benchmarks, we uniformly set the maximum generation length to \textbf{256}, the wavelet decomposition scale $B$ to \textbf{2}, and select the \textbf{Haar} wavelet as the base function. The threshold $\epsilon$ for confidence-based decoding is set to \textbf{0.9}. Crucially, the selection proportion of salient regions is dynamically adjusted within the range of \textbf{[0.3, 1]} across different datasets. Such highly consistent parameter settings across different models and benchmarks fully demonstrate the robustness of WaveFilter to hyperparameter variations.

To ensure a fair comparison, we standardized the configuration of experimental parameters. For the confidence-based decoding diffusion models, \textbf{LLaDA-8b-Instruct} and \textbf{Dream-v0-Base-7B}, the threshold is uniformly set to \textbf{0.9}. Meanwhile, when using Fast-dLLM, Elastic-Cache, and applying the WaveFilter framework, all hyperparameters strictly follow the \textbf{default settings} specified in their original papers. This approach aims to eliminate biases introduced by hyperparameter tuning, thereby objectively evaluating the native performance of each baseline method.

\begin{table*}[t]
\centering
\footnotesize
\caption{The hyper-parameters of WaveFilter under various benchmarks.}
\label{tab:hyperparameter}

\begin{tabular}{ccccccccc}
\toprule
\multirow{2}{*}{\textbf{Benchmark}} & \multirow{2}{*}{\textbf{Model}} & \multirow{2}{*}{\textbf{Generation length}} & \multirow{2}{*}{\textbf{Scale}} & \multirow{2}{*}{\textbf{Wavelet}} & \multicolumn{2}{c}{\textbf{Proportion}} & \multirow{2}{*}{\textbf{Threshold}} \\
\cmidrule(lr){6-7}
& & & & & $\boldsymbol{m_2}$ & $\boldsymbol{m_1}$ & \\
\midrule
\multirow{2}{*}{Longbench}     
& LLaDA-8B-Instruct & 256 & 2 & Haar & [0.3, 1] & [0.3, 1] & 0.9 \\ 
& Dream-v0-Base-7B  & 256 & 2 & Haar & [0.3, 1] & [0.3, 1] & 0.9 \\
\midrule
Ruler     
& LLaDA-8B-Instruct & 256 & 2 & Haar & [0.3, 1] & [0.3, 1] & 0.9 \\
\bottomrule
\end{tabular}
\end{table*}

\section{Token Statistics and Total Runtime}
\label{appendix:Performance}
To provide a transparent and rigorous assessment of the operational efficiency of the WaveFilter framework, this section details the comprehensive empirical statistics across various evaluation benchmarks. Table~\ref{tab:4} and Table~\ref{tab:5} thoroughly document the cumulative volume of generated tokens alongside the total execution time for LLaDA-8B-Instruct and Dream-v0-Base-7B across all subsets of the LongBench and Ruler benchmarks. Through an in-depth analysis of these experimental data, we not only gain profound insights into the efficiency of our framework but also reveal the inherent limitations of the throughput metric when evaluating long-text generation speeds, thereby elucidating the necessity of incorporating total runtime.

\begin{table*}
\caption{Comprehensive benchmark results of \textbf{LLaDA-8B-Instruct} and \textbf{Dream-v0-Base-7B} on \textbf{Longbench}. Each cell displays the \textbf{total generated tokens} (top row), and the \textbf{total runtime} in minutes (bottom row, \textcolor{green!60!black}{green: runtime}). The symbol "-" denotes \textbf{out of memory} errors.}
\label{tab:4}
\centering
\footnotesize 
\setlength{\tabcolsep}{3pt} 
\renewcommand{\arraystretch}{1.3} 
\resizebox{\textwidth}{!}{
\begin{tabular}{l cc ccc cc cc c}
\toprule
 & \multicolumn{2}{c}{\textbf{Single-Doc QA}} & \multicolumn{3}{c}{\textbf{Multi-Doc QA}} & \multicolumn{2}{c}{\textbf{Few-shot Learning}} & \multicolumn{2}{c}{\textbf{Synthetic}} & \multicolumn{1}{c}{\textbf{Code}} \\
\cmidrule(lr){2-3} \cmidrule(lr){4-6} \cmidrule(lr){7-8} \cmidrule(lr){9-10} \cmidrule(lr){11-11} 
& Qsp & MulF & HQA & 2WQA & MSQ & TREC & TrQA & PsgC & PsgR & LCC \\
\midrule
\textit{LLaDA-8B-Instruct} 
& \makecell{30104 \\ \scriptsize (\textcolor{green!60!black}{195.97})} 
& \makecell{29114 \\ \scriptsize (\textcolor{green!60!black}{238.75})} 
& \makecell{49658 \\ \scriptsize (\textcolor{green!60!black}{701.43})} 
& \makecell{46269 \\ \scriptsize (\textcolor{green!60!black}{385.63})} 
& \makecell{51157 \\ \scriptsize (\textcolor{green!60!black}{907.60})}
& \makecell{3733 \\ \scriptsize (\textcolor{green!60!black}{271.50})} 
& \makecell{5590 \\ \scriptsize (\textcolor{green!60!black}{591.47})}
& \makecell{50924 \\ \scriptsize (\textcolor{green!60!black}{461.50})} 
& \makecell{51119 \\ \scriptsize (\textcolor{green!60!black}{345.55})} 
& \makecell{125990 \\ \scriptsize (\textcolor{green!60!black}{310.86})}
\\
+ Fast-dLLM 
& \makecell{31211 \\ \scriptsize (\textcolor{green!60!black}{34.65})} 
& \makecell{29448 \\ \scriptsize (\textcolor{green!60!black}{42.35})} 
& \makecell{49867 \\ \scriptsize (\textcolor{green!60!black}{99.52})} 
& \makecell{46481 \\ \scriptsize (\textcolor{green!60!black}{60.17})} 
& \makecell{51176 \\ \scriptsize (\textcolor{green!60!black}{116.38})}
& \makecell{2459 \\ \scriptsize (\textcolor{green!60!black}{46.20})} 
& \makecell{7157 \\ \scriptsize (\textcolor{green!60!black}{90.77})}
& \makecell{50072 \\ \scriptsize (\textcolor{green!60!black}{100.52})} 
& \makecell{51096 \\ \scriptsize (\textcolor{green!60!black}{91.83})} 
& \makecell{124969 \\ \scriptsize (\textcolor{green!60!black}{69.02})}
\\

+ Fast-dLLM \& WaveFilter
& \makecell{\textbf{14806} \\ \scriptsize \colorbox{gray!20}{(\textcolor{green!60!black}{36.28})}}
& \makecell{\textbf{27226} \\ \scriptsize \colorbox{gray!20}{(\textcolor{green!60!black}{61.23})}} 
& \makecell{\textbf{49133} \\ \scriptsize \colorbox{gray!20}{(\textcolor{green!60!black}{143.47})}} 
& \makecell{\textbf{42826} \\ \scriptsize \colorbox{gray!20}{(\textcolor{green!60!black}{82.38})}} 
& \makecell{\textbf{51171} \\ \scriptsize \colorbox{gray!20}{(\textcolor{green!60!black}{159.23})}}
& \makecell{\textbf{1577} \\ \scriptsize \colorbox{gray!20}{(\textcolor{green!60!black}{67.07})}} 
& \makecell{\textbf{8873} \\ \scriptsize \colorbox{gray!20}{(\textcolor{green!60!black}{133.33})}}
& \makecell{\textbf{46346} \\ \scriptsize \colorbox{gray!20}{(\textcolor{green!60!black}{146.45})}} 
& \makecell{\textbf{51047} \\ \scriptsize \colorbox{gray!20}{(\textcolor{green!60!black}{138.78})}} 
& \makecell{\textbf{113990} \\ \scriptsize \colorbox{gray!20}{(\textcolor{green!60!black}{97.72})}}
\\

+ Elastic-Cache 
& \makecell{- \\ -} 
& \makecell{37878 \\ \scriptsize (\textcolor{green!60!black}{60.60})} 
& \makecell{- \\ -} 
& \makecell{- \\ -} 
& \makecell{- \\ -}
& \makecell{4001 \\ \scriptsize (\textcolor{green!60!black}{60.60})} 
& \makecell{- \\ -}
& \makecell{- \\ -} 
& \makecell{51087 \\ \scriptsize (\textcolor{green!60!black}{159.93})} 
& \makecell{- \\ -}
\\

+  Elastic-Cache \& WaveFilter
& \makecell{- \\ -} 
& \makecell{\textbf{37491} \\ \scriptsize \colorbox{gray!20}{(\textcolor{green!60!black}{57.45})}} 
& \makecell{- \\ -} 
& \makecell{- \\ -} 
& \makecell{- \\ -}
& \makecell{\textbf{4719} \\ \scriptsize \colorbox{gray!20}{(\textcolor{green!60!black}{107.07})}} 
& \makecell{- \\ -}
& \makecell{- \\ -} 
& \makecell{\textbf{51097} \\ \scriptsize \colorbox{gray!20}{(\textcolor{green!60!black}{166.47})}} 
& \makecell{- \\ -}

\\
\midrule

\textit{Dream-v0-Base-7B} 
& \makecell{50985 \\ \scriptsize (\textcolor{green!60!black}{244.77})} 
& \makecell{38212 \\ \scriptsize (\textcolor{green!60!black}{288.47})} 
& \makecell{51125 \\ \scriptsize (\textcolor{green!60!black}{773.72})} 
& \makecell{51115 \\ \scriptsize (\textcolor{green!60!black}{371.90})} 
& \makecell{51152 \\ \scriptsize (\textcolor{green!60!black}{986.82})}
& \makecell{51200 \\ \scriptsize (\textcolor{green!60!black}{264.58})} 
& \makecell{51200 \\ \scriptsize (\textcolor{green!60!black}{624.38})}
& \makecell{51198 \\ \scriptsize (\textcolor{green!60!black}{387.15})} 
& \makecell{51200 \\ \scriptsize (\textcolor{green!60!black}{297.30})} 
& \makecell{127986 \\ \scriptsize (\textcolor{green!60!black}{226.57})}
\\

+ Fast-dLLM 
& \makecell{51003 \\ \scriptsize (\textcolor{green!60!black}{30.30})} 
& \makecell{38163 \\ \scriptsize (\textcolor{green!60!black}{29.72})} 
& \makecell{51106 \\ \scriptsize (\textcolor{green!60!black}{66.36})} 
& \makecell{51115 \\ \scriptsize (\textcolor{green!60!black}{38.78})} 
& \makecell{51123 \\ \scriptsize (\textcolor{green!60!black}{79.70})}
& \makecell{51200 \\ \scriptsize (\textcolor{green!60!black}{33.92})} 
& \makecell{51199 \\ \scriptsize (\textcolor{green!60!black}{62.23})} 
& \makecell{51192 \\ \scriptsize (\textcolor{green!60!black}{60.58})}
& \makecell{51196 \\ \scriptsize (\textcolor{green!60!black}{51.07})} 
& \makecell{127973 \\ \scriptsize (\textcolor{green!60!black}{47.22})}
\\

+ Fast-dLLM \& WaveFilter
& \makecell{\textbf{50962} \\ \scriptsize \colorbox{gray!20}{(\textcolor{green!60!black}{50.98})}} 
& \makecell{\textbf{38181} \\ \scriptsize \colorbox{gray!20}{(\textcolor{green!60!black}{55.90})}} 
& \makecell{\textbf{51064} \\ \scriptsize \colorbox{gray!20}{(\textcolor{green!60!black}{112.65})}} 
& \makecell{\textbf{51129} \\ \scriptsize \colorbox{gray!20}{(\textcolor{green!60!black}{63.87})}} 
& \makecell{\textbf{51050} \\ \scriptsize \colorbox{gray!20}{(\textcolor{green!60!black}{140.33})}}
& \makecell{\textbf{51200} \\ \scriptsize \colorbox{gray!20}{(\textcolor{green!60!black}{66.78})}} 
& \makecell{\textbf{51200} \\ \scriptsize \colorbox{gray!20}{(\textcolor{green!60!black}{115.03})}}
& \makecell{\textbf{51140} \\ \scriptsize \colorbox{gray!20}{(\textcolor{green!60!black}{102.42})}} 
& \makecell{\textbf{51176} \\ \scriptsize \colorbox{gray!20}{(\textcolor{green!60!black}{84.58})}} 
& \makecell{\textbf{127960} \\ \scriptsize \colorbox{gray!20}{(\textcolor{green!60!black}{74.38})}}
\\

+ Elastic-Cache 
& \makecell{- \\ -} 
& \makecell{38282 \\ \scriptsize (\textcolor{green!60!black}{20.37})} 
& \makecell{- \\ -} 
& \makecell{51162 \\ \scriptsize (\textcolor{green!60!black}{61.07})} 
& \makecell{- \\ -}
& \makecell{51200 \\ \scriptsize (\textcolor{green!60!black}{81.02})} 
& \makecell{- \\ -}
& \makecell{- \\ -} 
& \makecell{51199 \\ \scriptsize (\textcolor{green!60!black}{62.13})} 
& \makecell{- \\ -}
\\

+  Elastic-Cache \& WaveFilter
& \makecell{- \\ -} 
& \makecell{\textbf{38271} \\ \scriptsize \colorbox{gray!20}{(\textcolor{green!60!black}{16.88})}} 
& \makecell{- \\ -} 
& \makecell{\textbf{51108} \\ \scriptsize \colorbox{gray!20}{(\textcolor{green!60!black}{69.17})}} 
& \makecell{- \\ -}
& \makecell{\textbf{51120} \\ \scriptsize \colorbox{gray!20}{(\textcolor{green!60!black}{77.43})}} 
& \makecell{- \\ -}
& \makecell{- \\ -} 
& \makecell{\textbf{51195} \\ \scriptsize \colorbox{gray!20}{(\textcolor{green!60!black}{65.75})}} 
& \makecell{- \\ -}
\\
\bottomrule
\end{tabular}}
\end{table*}

\begin{table*}
\caption{Comprehensive benchmark results of \textbf{LLaDA-8B-Instruct} on \textbf{Ruler}. 
Each cell displays the \textbf{total generated tokens} (top row), and the \textbf{total runtime} in minutes (bottom row, \textcolor{green!60!black}{green: runtime}).}
\label{tab:5}
\centering
\footnotesize
\resizebox{\textwidth}{!}{
\begin{tabular}{l | cccccccccc | c}
\toprule
 & s1 & s2 & m1 & m2 & mv & mq & vt & cwe & qa1 & qa2 & Context Length \\
\midrule
\textit{LLaDA-8B-Instruct} 
& \makecell{15049 \\ \scriptsize (\textcolor{green!60!black}{65.97})}
& \makecell{61358 \\ \scriptsize (\textcolor{green!60!black}{125.97})}
& \makecell{43153 \\ \scriptsize (\textcolor{green!60!black}{98.18})}
& \makecell{116706 \\ \scriptsize (\textcolor{green!60!black}{390.07})}
& \makecell{127503 \\ \scriptsize (\textcolor{green!60!black}{460.52})}
& \makecell{127486 \\ \scriptsize (\textcolor{green!60!black}{334.87})}
& \makecell{118877 \\ \scriptsize (\textcolor{green!60!black}{308.05})}
& \makecell{127538 \\ \scriptsize (\textcolor{green!60!black}{334.60})}
& \makecell{16429 \\ \scriptsize (\textcolor{green!60!black}{64.45})}
& \makecell{36839 \\ \scriptsize (\textcolor{green!60!black}{116.90})}
& 4K 
\\
+ Fast-dLLM 
& \makecell{53474 \\ \scriptsize (\textcolor{green!60!black}{39.68})}
& \makecell{97647 \\ \scriptsize (\textcolor{green!60!black}{75.75})}
& \makecell{81597 \\ \scriptsize (\textcolor{green!60!black}{63.55})}
& \makecell{127256 \\ \scriptsize (\textcolor{green!60!black}{92.40})}
& \makecell{127501 \\ \scriptsize (\textcolor{green!60!black}{71.78})}
& \makecell{127486 \\ \scriptsize (\textcolor{green!60!black}{67.60})}
& \makecell{125999 \\ \scriptsize (\textcolor{green!60!black}{73.88})}
& \makecell{127598 \\ \scriptsize (\textcolor{green!60!black}{69.92})}
& \makecell{23191 \\ \scriptsize (\textcolor{green!60!black}{31.98})}
& \makecell{49151 \\ \scriptsize (\textcolor{green!60!black}{47.12})}
& 4K
\\
+ Fast-dLLM  \& WaveFilter
& \makecell{\textbf{11846} \\ \scriptsize \colorbox{gray!20}{(\textcolor{green!60!black}{33.20})}}
& \makecell{\textbf{12360} \\ \scriptsize \colorbox{gray!20}{(\textcolor{green!60!black}{33.48})}}
& \makecell{\textbf{11283} \\ \scriptsize \colorbox{gray!20}{(\textcolor{green!60!black}{32.38})}}
& \makecell{\textbf{127015} \\ \scriptsize \colorbox{gray!20}{(\textcolor{green!60!black}{152.13})}}
& \makecell{\textbf{90481} \\ \scriptsize \colorbox{gray!20}{(\textcolor{green!60!black}{80.97})}}
& \makecell{\textbf{124450} \\ \scriptsize \colorbox{gray!20}{(\textcolor{green!60!black}{117.52})}}
& \makecell{\textbf{107579} \\ \scriptsize \colorbox{gray!20}{(\textcolor{green!60!black}{83.48})}}
& \makecell{\textbf{114729} \\ \scriptsize \colorbox{gray!20}{(\textcolor{green!60!black}{118.32})}}
& \makecell{\textbf{11470} \\ \scriptsize \colorbox{gray!20}{(\textcolor{green!60!black}{29.07})}}
& \makecell{\textbf{43783} \\ \scriptsize \colorbox{gray!20}{(\textcolor{green!60!black}{61.20})}}
& 4K
\\
+ Elastic-Cache 
& \makecell{123382 \\ \scriptsize (\textcolor{green!60!black}{31.50})}
& \makecell{126300 \\ \scriptsize (\textcolor{green!60!black}{97.77})}
& \makecell{125387 \\ \scriptsize (\textcolor{green!60!black}{70.27})}
& \makecell{127493 \\ \scriptsize (\textcolor{green!60!black}{122.20})}
& \makecell{127493 \\ \scriptsize (\textcolor{green!60!black}{113.32})}
& \makecell{127471 \\ \scriptsize (\textcolor{green!60!black}{85.95})}
& \makecell{127142 \\ \scriptsize (\textcolor{green!60!black}{95.42})}
& \makecell{127614 \\ \scriptsize (\textcolor{green!60!black}{235.60})}
& \makecell{123317 \\ \scriptsize (\textcolor{green!60!black}{22.22})}
& \makecell{124669 \\ \scriptsize (\textcolor{green!60!black}{45.40})}
& 4K
\\
+ Elastic-Cache \& WaveFilter
& \makecell{\textbf{123135} \\ \scriptsize \colorbox{gray!20}{(\textcolor{green!60!black}{37.72})}}
& \makecell{\textbf{126118} \\ \scriptsize \colorbox{gray!20}{(\textcolor{green!60!black}{117.70})}}
& \makecell{\textbf{125751} \\ \scriptsize \colorbox{gray!20}{(\textcolor{green!60!black}{73.55})}}
& \makecell{\textbf{127481} \\ \scriptsize \colorbox{gray!20}{(\textcolor{green!60!black}{140.58})}}
& \makecell{\textbf{127497} \\ \scriptsize \colorbox{gray!20}{(\textcolor{green!60!black}{124.92})}}
& \makecell{\textbf{127463} \\ \scriptsize \colorbox{gray!20}{(\textcolor{green!60!black}{103.67})}}
& \makecell{\textbf{127097} \\ \scriptsize \colorbox{gray!20}{(\textcolor{green!60!black}{111.20})}}
& \makecell{\textbf{127580} \\ \scriptsize \colorbox{gray!20}{(\textcolor{green!60!black}{285.80})}}
& \makecell{\textbf{123255} \\ \scriptsize \colorbox{gray!20}{(\textcolor{green!60!black}{23.30})}}
& \makecell{\textbf{124647} \\ \scriptsize \colorbox{gray!20}{(\textcolor{green!60!black}{48.07})}}
& 4K
\\

\midrule
\textit{LLaDA-8B-Instruct} 
& \makecell{90628 \\ \scriptsize (\textcolor{green!60!black}{370.22})}
& \makecell{127543 \\ \scriptsize (\textcolor{green!60!black}{1227.47})}
& \makecell{127553 \\ \scriptsize (\textcolor{green!60!black}{1021.02})}
& \makecell{127500 \\ \scriptsize (\textcolor{green!60!black}{1111.00})}
& \makecell{127560 \\ \scriptsize (\textcolor{green!60!black}{690.32})}
& \makecell{127600 \\ \scriptsize (\textcolor{green!60!black}{784.80})}
& \makecell{126536 \\ \scriptsize (\textcolor{green!60!black}{671.53})}
& \makecell{127897 \\ \scriptsize (\textcolor{green!60!black}{759.45})}
& \makecell{127499 \\ \scriptsize (\textcolor{green!60!black}{802.55})}
& \makecell{127048 \\ \scriptsize (\textcolor{green!60!black}{886.93})}
& 8K
\\
+ Fast-dLLM 
& \makecell{110022 \\ \scriptsize (\textcolor{green!60!black}{107.43})}
& \makecell{127524 \\ \scriptsize (\textcolor{green!60!black}{185.78})}
& \makecell{127531 \\ \scriptsize (\textcolor{green!60!black}{164.82})}
& \makecell{127495 \\ \scriptsize (\textcolor{green!60!black}{184.23})}
& \makecell{127488 \\ \scriptsize (\textcolor{green!60!black}{106.98})}
& \makecell{127610 \\ \scriptsize (\textcolor{green!60!black}{118.70})}
& \makecell{126509 \\ \scriptsize (\textcolor{green!60!black}{119.92})}
& \makecell{127996 \\ \scriptsize (\textcolor{green!60!black}{155.02})}
& \makecell{127500 \\ \scriptsize (\textcolor{green!60!black}{137.07})}
& \makecell{127242 \\ \scriptsize (\textcolor{green!60!black}{154.78})}
& 8K
\\
+ Fast-dLLM \& WaveFilter
& \makecell{\textbf{22661} \\ \scriptsize \colorbox{gray!20}{(\textcolor{green!60!black}{105.58})}}
& \makecell{\textbf{125781} \\ \scriptsize \colorbox{gray!20}{(\textcolor{green!60!black}{273.08})}}
& \makecell{\textbf{127260} \\ \scriptsize \colorbox{gray!20}{(\textcolor{green!60!black}{245.25})}}
& \makecell{\textbf{121832} \\ \scriptsize \colorbox{gray!20}{(\textcolor{green!60!black}{288.02})}}
& \makecell{\textbf{127681} \\ \scriptsize \colorbox{gray!20}{(\textcolor{green!60!black}{161.70})}}
& \makecell{\textbf{127807} \\ \scriptsize \colorbox{gray!20}{(\textcolor{green!60!black}{175.17})}}
& \makecell{\textbf{126695} \\ \scriptsize \colorbox{gray!20}{(\textcolor{green!60!black}{161.07})}}
& \makecell{\textbf{127002} \\ \scriptsize \colorbox{gray!20}{(\textcolor{green!60!black}{218.22})}}
& \makecell{\textbf{127546} \\ \scriptsize \colorbox{gray!20}{(\textcolor{green!60!black}{198.66})}}
& \makecell{\textbf{120473} \\ \scriptsize \colorbox{gray!20}{(\textcolor{green!60!black}{218.01})}}

& 8K
\\
+ Elastic-Cache 
& \makecell{127493 \\ \scriptsize (\textcolor{green!60!black}{171.53})}
& \makecell{127556 \\ \scriptsize (\textcolor{green!60!black}{303.01})}
& \makecell{127586 \\ \scriptsize (\textcolor{green!60!black}{268.70})}
& \makecell{127402 \\ \scriptsize (\textcolor{green!60!black}{281.40})}
& \makecell{127469 \\ \scriptsize (\textcolor{green!60!black}{210.05})}
& \makecell{127697 \\ \scriptsize (\textcolor{green!60!black}{197.65})}
& \makecell{127413 \\ \scriptsize (\textcolor{green!60!black}{183.58})}
& \makecell{127910 \\ \scriptsize (\textcolor{green!60!black}{602.37})}
& \makecell{127491 \\ \scriptsize (\textcolor{green!60!black}{205.03})}
& \makecell{127499 \\ \scriptsize (\textcolor{green!60!black}{206.60})}
& 8K
\\
+ Elastic-Cache \& WaveFilter
& \makecell{\textbf{127339} \\ \scriptsize \colorbox{gray!20}{(\textcolor{green!60!black}{229.25})}} 
& \makecell{\textbf{127570} \\ \scriptsize \colorbox{gray!20}{(\textcolor{green!60!black}{286.17})}} 
& \makecell{\textbf{127618} \\ \scriptsize \colorbox{gray!20}{(\textcolor{green!60!black}{237.87})}} 
& \makecell{\textbf{127453} \\ \scriptsize \colorbox{gray!20}{(\textcolor{green!60!black}{378.28})}} 
& \makecell{\textbf{127544} \\ \scriptsize \colorbox{gray!20}{(\textcolor{green!60!black}{210.50})}} 
& \makecell{\textbf{127623} \\ \scriptsize \colorbox{gray!20}{(\textcolor{green!60!black}{210.95})}} 
& \makecell{\textbf{127347} \\ \scriptsize \colorbox{gray!20}{(\textcolor{green!60!black}{236.15})}} 
& \makecell{\textbf{127931} \\ \scriptsize \colorbox{gray!20}{(\textcolor{green!60!black}{897.65})}} 
& \makecell{\textbf{127492} \\ \scriptsize \colorbox{gray!20}{(\textcolor{green!60!black}{218.72})}} 
& \makecell{\textbf{127497} \\ \scriptsize \colorbox{gray!20}{(\textcolor{green!60!black}{252.97})}} 
& 8K
\\
\midrule
\end{tabular}
}
\end{table*}

\section{Generation Examples}
\label{append:example}
Appendix~\ref{box1} and Appendix~\ref{box2} present the output snippets of Fast-dLLM and Fast-dLLM integrated with the WaveFilter framework, respectively, on the 4K context length s1 subset of the Ruler benchmark. A comparative analysis reveals that the generation from Fast-dLLM contains a significant amount of redundant information irrelevant to the prompt, whereas the introduction of WaveFilter enables the model to accurately generate highly relevant answers. It is worth noting that in Table~\ref{tab:1} and Table~\ref{tab:2}, although Fast-dLLM achieves higher throughput, it contains invalid generations. This phenomenon demonstrates that in long-context tasks, relying solely on throughput as a metric fails to objectively reflect the actual operational efficiency and generation quality of the model. This clear and illustrative comparative example compellingly demonstrates that for complex long-context tasks, throughput should be integrated with total execution time and final generation quality to establish a comprehensive, multi-dimensional evaluation framework.

\subsection{Case Study: Fast-dLLM}
\begin{tcolorbox}[
    breakable,
    title=Examples of Fast-dLLM,
    fontupper=\fontsize{8pt}{12.9pt}\selectfont,
    label=box1,
]
\columnseprule=0.5pt
\sloppy

\textbf{"Context":} "A special magic number is hidden within the following text. Make sure to memorize it.
The grass is green. The sky is blue. The sun is yellow. Here we go. There and back again. \\
\indent \dots \textit{[Long context text omitted]} \dots \\
\textbf{One of the special magic numbers for capable-percent is: 7374122.} \\
\indent \dots \textit{[Long context text omitted]} \dots \\
The grass is green. The sky is blue. The sun is yellow. Here we go. There and back again."

\textbf{"Question":} "What is the special magic number for wandering-age mentioned in the provided text? The special magic number for wandering-age mentioned in the provided text is"

\textbf{"Answer":} "The special magic number for capable-percent mentioned in the provided text is 7374122.

The grass is green. The sky is blue. The sun is yellow. Here we go. There and back again.
The grass is green. The sky is blue. The sun is yellow. Here we go. There and back again.
The grass is green. The sky is blue. The sun is yellow. Here we go. There and back again.
The grass is green. The sky is blue. The sun is yellow. Here we go. There and back again.
The grass is green. The sky is blue. The sun is yellow. Here we go. There and back again.
The grass is green. The sky is blue. The sun is yellow. Here we go. There and back again.
The grass is green. The sky is blue. The sun is yellow. Here we go. There and back again.
The grass is green. The sky is blue. The sun is yellow. Here we go. There and back again.
The grass is green. The sky is blue. The sun is yellow. Here we go. There and back again.
"
\end{tcolorbox}

\subsection{Case Study: Fast-dLLM \& WaveFilter}
\begin{tcolorbox}[
    breakable,
    title=Examples of Fast-dLLM \& WaveFilter,
    fontupper=\fontsize{8pt}{13pt}\selectfont,
    label=box2,
]
\columnseprule=0.5pt
\sloppy

\textbf{"Context":} "A special magic number is hidden within the following text. Make sure to memorize it.
The grass is green. The sky is blue. The sun is yellow. Here we go. There and back again. \\
\indent \dots \textit{[Long context text omitted]} \dots \\
\textbf{One of the special magic numbers for capable-percent is: 7374122.} \\
\indent \dots \textit{[Long context text omitted]} \dots \\
The grass is green. The sky is blue. The sun is yellow. Here we go. There and back again."

\textbf{"Question":} "What is the special magic number for wandering-age mentioned in the provided text? The special magic number for wandering-age mentioned in the provided text is"

\textbf{"Answer":} "The special magic number for capable-percent mentioned in the provided text is 7374122."
\end{tcolorbox}

\clearpage
\section{The Use of Large Language Models}
During the preparation of this paper, the use of Large Language Models was strictly limited to language polishing and text translation. All core scientific contributions—including research conceptualization, experimental design and execution, data analysis, and conclusion derivation—were independently completed by the authors, who assume full responsibility for this work. Following a rigorous verification, the final text has been ensured to be accurate, factually correct, and in full compliance with academic integrity standards.
\end{document}